\documentclass{vldb}
\usepackage{balance}

\usepackage{times}
\usepackage{graphicx} 
\usepackage{subfigure}
\usepackage{color}
\usepackage{caption}
\usepackage{float}

\usepackage[usenames,dvipsnames]{xcolor}

\usepackage{algorithm}
\usepackage{algpseudocode}
\usepackage{amsmath}
\usepackage{amsfonts}
\usepackage{enumitem}
\usepackage{multirow}
\usepackage{listings}
\usepackage[rightcaption]{sidecap}

\usepackage{bm}
\usepackage{multicol,blindtext}
\newcommand{\vect}[1]{\boldsymbol{\mathbf{#1}}}

\newcommand{\row}{row}
\newcommand{\col}{col}

\newcommand\sys[1]{$#1$}

\newcommand{\argmax}{\operatornamewithlimits{argmax}}

\usepackage{url}

\newcommand{\alphav}{\vect\alpha}

\newcommand{\phiv}{\vect\phi}

\newif\ifdebug
\debugtrue

\ifdebug
\newcommand{\chrisx}[1]{{\color{red}{\bf\sf [CJF: #1]}}}
\newcommand{\junz}[1]{{\color{blue}{\bf\sf [ZJ: #1]}}}
\newcommand{\arnie}[1]{{\color{cyan}{\bf\sf [AB: #1]}}}
\newcommand{\lkw}[1]{{\color{purple}{\bf\sf [LKW: #1]}}}
\newcommand{\cwg}[1]{{\color{green}{\bf\sf [CWG: #1]}}}
\else
\newcommand{\chrisx}[1]{{\color{red}{}}}
\newcommand{\junz}[1]{{\color{blue}{}}}
\newcommand{\arnie}[1]{{\color{cyan}{}}}
\newcommand{\lkw}[1]{{\color{purple}{}}}
\newcommand{\cwg}[1]{{\color{green}{}}}
\fi

\begin{document}

\lstset{
    basicstyle=\tt,
    keywordstyle=\color{blue!70}, commentstyle=\color{red!50!green!50!blue!50},
    frame=shadowbox,
    rulesepcolor=\color{red!20!green!20!blue!20}
}

\title{WarpLDA: a Cache Efficient O(1) Algorithm for \\
Latent Dirichlet Allocation}

\numberofauthors{1} 
%
\author{
%
%
\alignauthor
Jianfei Chen$^{\dag \ddag}$, Kaiwei Li$^{\dag\S}$, Jun Zhu$^{\dag\ddag}$, Wenguang Chen$^{\dag\S}$
\\
      \affaddr{$^\dag$Dept. of Comp. Sci. \& Tech.; TNList Lab; CBICR Center; Tsinghua University}\\
\affaddr{$^\ddag$State Key Lab of Intelligent Technology \& Systems, Beijing, 100084 China}\\
      \affaddr{$^\S$Parallel Architecture \& Compiler Tech. of Mobile, Accelerated, and Networked Systems (PACMAN)}\\
       \email{\{chenjian14, likw14\}@mails.tsinghua.edu.cn;\{dcszj,~cwg\}@tsinghua.edu.cn}
}

\maketitle

\begin{abstract}
Developing efficient and scalable algorithms for Latent Dirichlet Allocation (LDA) is of wide interest for many applications. Previous work has developed an $O(1)$ Metropolis-Hastings (MH) sampling method for each token. However, the performance is far from being optimal due to frequent \emph{cache misses} caused by random accesses to the parameter matrices. 

In this paper, we first carefully analyze the memory access behaviour of existing algorithms for LDA by cache locality at  document level.
We then develop WarpLDA, which achieves  $O(1)$ time complexity per-token and fits the randomly accessed memory per-document in the L3 cache. Our empirical results in a wide range of testing conditions demonstrate that WarpLDA is consistently 5-15x faster than the state-of-the-art MH-based LightLDA, and is faster than the state-of-the-art sparsity aware F+LDA in most settings. WarpLDA manages to learn up to one million topics from billion-scale documents  
in 5 hours, at an unprecedentedly throughput of 11G tokens per second.
\end{abstract}

\section{Introduction}\label{sec:introduction}
Topic modeling provides a suite of statistical tools to discover latent semantic structures from complex corpora, with latent Dirichlet allocation (LDA)~\cite{blei2003latent} as the most popular one. LDA has found many applications in text analysis~\cite{boyd2007topic,zhu12jmlr}, data visualization~\cite{iwata2008probabilistic,liu2014topicpanorama}, recommendation systems~\cite{chen2009collaborative}, information retrieval~\cite{wei2006lda} and network analysis~\cite{Chang:RTM09,chen2015rtm}. LDA represents each document as an admixture of topics, each of which is a unigram distribution of words. Since exact inference is intractable, both variational Bayes (VB) and Markov Chain Monte Carlo (MCMC) methods have been developed for approximate inference, including mean-field variational Bayes~\cite{blei2003latent}, collapsed variational Bayes~\cite{teh2006collapsed}, collapsed Gibbs sampling (CGS)~\cite{griffiths2004finding} and expectation propagation~\cite{minka2002expectation}. Among these methods, CGS is most popular due to its simplicity and availability for fast sparsity-aware algorithms~\cite{yao2009efficient,li2014reducing,yuan2014lightlda}.

Entering the Big Data era, applications often require \emph{large-scale} topic modeling to boost their performance. For example, Wang et al.~\cite{wang2014peacock,yuan2014lightlda} show that learning 1 million topics can lead to significant performance gain on various tasks such as advertisement and recommendation. Other recent endeavours on learning large topic models often contain billions of documents, millions of topics, and millions of unique tokens~\cite{wang2014peacock,yuan2014lightlda,li2014scaling}. 
Various fast sampling algorithms have been proposed for LDA, reducing the time complexity of sampling the token assignment per-token from $O(K)$ to $O(1)$, where $K$ is the number of topics~\cite{yao2009efficient, li2014scaling, yuan2014lightlda}. 

Although many efforts have been spent on improving the per-token sampling complexity, little attention has been paid to examine the cache locality, another important dimension to improve the overall efficiency. For all the aforementioned algorithms, 
the cache locality 
is not getting better; in fact some are even getting \emph{worse} (See Table~\ref{tbl:summary-of-algorithms} for details).
The running time of LDA is often dominated by \emph{random memory accesses}, where the time consumption is roughly proportional to the latency of each access. 
As shown in Table~\ref{tbl:cache-latency}, the latency of accessing different levels of the memory hierarchy varies greatly, and accessing a higher-level cache can be orders of magnitude faster than accessing a lower-level cache or the main memory. 
As we will show in Sec.~\ref{sec:analysis}, when processing a single document, the random accesses of previous LDA algorithms spread across either an $O(KV)$ matrix or an $O(DK)$ matrix, where $D$ is the number of documents and $V$ is the vocabulary size. As $K$, $V$ and $D$ can all exceed one million in large-scale applications, the matrix can be tens of gigabytes in size, which is too large to fit in any cache, resulting in unsatisfactory memory efficiency. Moreover, this size is difficult to reduce for CGS because both the document-wise counts and the word-wise counts need to be accessed for sampling a single token.

\begin{table}[t]
\caption{Configuration of the memory hierarchy in an Intel Ivy Bridge CPU. L1D denotes L1 data cache, and $^*$ stands for per core. \label{tbl:cache-latency}}\vspace{-.4cm}
\centering
\begin{tabular}{lllll}
\hline
            & L1D & L2 & L3 & Main memory \\ \hline
Latency (cycles) & 5 & 12 & 30 & 180+ \\ 
Size & 32KB$^{*}$ & 256KB$^*$ & 30MB & 10GB+ \\
\hline
\end{tabular}\vspace{-.4cm}
\end{table}


In this paper, we propose to reduce the latency of random accesses of LDA by reducing the size of the randomly accessed memory. 
Based on a careful analysis of existing algorithms, 
we develop WarpLDA\footnote{The name comes after Warp Drive, the hypothetical faster-than-light propulsion system in Star Trek.}, a novel sampling algorithm based on Monte-Carlo Expectation Maximization (MCEM) that preserves the best $O(1)$ time complexity per-token and has some carefully designed reordering strategy to achieve an $O(K)$ size of randomly accessed memory per-document, which is small enough to fit in the L3 cache. As the L3 cache is at least six times faster than the main memory (See Table~\ref{tbl:cache-latency}), this could result in a significant performance gain. Another nice property of WarpLDA is that it simplifies the system design. We present an implementation of WarpLDA in a distributed and memory-efficient way, with a carefully designed computational framework based on distributed sparse matrices. 

Extensive empirical studies in a wide range of settings demonstrate that WarpLDA consistently converges 5-15x faster than the state-of-the-art Metropolis-Hastings based LightLDA~\cite{yuan2014lightlda} and in most settings, faster than the state-of-the-art sparsity aware F+LDA~\cite{yu2015scalable} for learning LDA models with thousands to hundreds of thousands topics. WarpLDA scales to corpora of billion-scale documents, and achieves an unprecedentedly 11G tokens per second throughput with 256 machines.

\textbf{Outline: } Section 2 introduces some basics of LDA, Section 3 provides the memory efficiency analysis of existing algorithms. Section 4 introduces the WarpLDA algorithm, and Section 5 provides system design and implementation details. Experiments are presented in Section 6. Section 7 concludes.

\begin{table*}[t]
\centering
\caption{Summary of existing algorithms of LDA, where $K$: number of topics, $K_d$: average number of topics per-document, $K_w$: average number of topics per word, $D$: number of documents, $V$: size of vocabulary, SA: sparsity-aware algorithm, MH: MH-based algorithm.
\label{tbl:summary-of-algorithms}}
\vspace{-.2cm}
\setlength\tabcolsep{2pt}

\begin{tabular}{cccccccc}
\hline
Algorithm & Type &  Amount of sequential accesses & Number of random accesses & Size of randomly accessed memory & Order \\
         & & $O$(per-token) & $O$(per-token) &per-document &  \\
\hline
CGS~\cite{griffiths2004finding} & - & $K$  & -& - & doc  \\
SparseLDA~\cite{yao2009efficient} & SA & $K_d+K_w$ & $K_d+K_w$ & $KV$ & doc  \\
AliasLDA~\cite{li2014reducing} & SA\&MH  & $K_d$ & $K_d$ & $KV$ & doc   \\
F+LDA~\cite{yu2015scalable} & SA & $K_d$ & $K_d$ & $DK$ & word   \\
LightLDA~\cite{yuan2014lightlda} & MH & - & $1$ & $KV$ & doc  \\
\hline
\textbf{WarpLDA} & MH &  - & \textbf{1} & $\boldsymbol{K}$ & doc\&word   \\
\hline
\end{tabular}
\vspace{-.2cm}

\end{table*}



\section{Backgrounds}
\subsection{Basics of LDA}\label{sec:basics-of-lda}
Let $\vect w_d = \{w_{dn}\}_{n=1}^{L_d}$ denote the collection of $L_d$ words in document $d$ and $\vect W = \{\vect w_{d}\}_{d=1}^D$ be a collection of $D$ documents. Let $V$ denote the vocabulary size. 
Latent Dirichlet Allocation~\cite{blei2003latent} is a hierarchical Bayesian model, which models the distribution of a word as a mixture of $K$ topic distributions, with a shared mixing proportion for words within the same document. 
Formally, each topic $k$ is a $V$-dim word distribution $\vect\phi_k$, which follows a Dirichlet prior $\vect\phi_k \sim \mbox{Dir}(\beta\vect 1)$ with parameter $\beta$; and the generative process of LDA for each document $d$ is:
\begin{itemize}[label={}]
  \setlength\itemsep{0em}
  \item Draw a $K$-dim topic mixing proportion: $\vect\theta_d \sim \mbox{Dir}(\vect \alpha)$,
    \item For each position $n \in \{1, \dots, L_d\}$:
    \begin{itemize}[label={}]
        \setlength\itemsep{0em}
        \item Draw a topic assignment: $z_{dn} \sim \mbox{Mult}( \vect\theta_d)$,
        \item Draw a word: $w_{dn} \sim \mbox{Mult}( \vect\phi_{z_{dn}})$,
    \end{itemize}
\end{itemize}
where $\mbox{Mult}(\cdot)$ is a multinomial distribution (or categorical distribution), and 
$\vect\alpha$ are Dirichlet parameters.  
Let $\vect\Phi = [\phiv_1 \cdots \phiv_K]$ be the $K \times V$ topic matrix. We further denote $\vect\Theta = \{\vect\theta_d \}_{d=1}^D$ and $\vect Z = \{\vect z_d\}_{d=1}^D$, where $\vect z_d = \{z_{dn}\}_{n=1}^{L_d}$. Let $\bar\alpha = \sum_{k=1}^K  \alpha_k$ and $\bar\beta = V\beta$. In this paper, we define \emph{token} as an occurrence of a \emph{word}, e.g., ``apple'' is a word, and each of its occurrence is a token. When we say ``all tokens of word $w$'' we mean all the occurrences of $w$, which may have different topic assignments and we use $\vect z_w$ to denote the topic assignments of all tokens of word $w$. 

To train LDA, one must infer the posterior distribution (or its marginal version) of latent variables $(\vect\Theta, \vect\Phi, \vect Z)$ given $(\vect W, \vect \alpha, \beta)$. Unfortunately, exact posterior inference is intractable. Thus approximate techniques including variational Bayes and Markov Chain Monte Carlo (MCMC) methods are adopted. As mentioned before, Collapsed Gibbs Sampling (CGS)~\cite{griffiths2004finding} is most popular because of its simplicity and the availability of fast sampling algorithms. Given $(\vect W, \vect\alpha, \beta)$, CGS integrates out $(\vect\Theta, \vect\Phi)$ by conjugacy and iteratively samples $z_{dn}$ from the local conditional distribution:
\setlength\arraycolsep{-3pt} \begin{eqnarray}
&& p(z_{dn} \!\!=\! k | \vect Z_{\neg dn}, w_{dn} \!\!=\! w, \vect W_{\neg dn}) 
\propto (C_{dk}^{\neg dn} \!+\! \alpha_k)\frac{C_{wk}^{\neg dn} \!+\! \beta}{C_k^{\neg dn} \!+\! \bar\beta},\label{eqn:cgs}
\end{eqnarray}
where $C_{dk}=\sum_{n=1}^{L_d} \mathbb{I}(z_{dn}=k)$ is the number of tokens that are assigned to topic $k$ in document $d$; $C_{wk}=\sum_{d=1}^D \sum_{n=1}^{L_d}$ $\mathbb{I}(z_{dn}=k,w_{dn}=w)$ is the number of times that word $w$ has topic $k$; $C_k = \sum_d C_{dk} = \sum_w C_{wk}$. \footnote{
We distinguish different counts by their subscripts.} The superscript or subscript $^{\neg dn}$ stands for excluding ($z_{dn}, w_{dn}$) from the corresponding count or collection.  We further define $\vect C_d$ to be the $D\times K$ matrix formed by $C_{dk}$, and $\vect C_w$ to be the $V\times K$ matrix formed by $C_{wk}$, with $\vect c_d$ and $\vect c_w$ being their particular rows indexed by $d$ and $w$, and $K_d$, $K_w$ being the number of non-zero entries of the corresponding row. Let the global topic count vector $\vect c_k = (C_1, \dots, C_K)^\top$. 

By sampling in a collapsed space, CGS often converges faster than a standard Gibbs sampler in a space with all the variables. 
A straightforward implementation of Eq.~(\ref{eqn:cgs}) is of complexity $O(K)$ per-token by naively enumerating all the $K$ possible topic assignments. This can be too expensive for large-scale applications where $K$ can be in the order of $10^6$. Various fast sampling algorithms~\cite{yao2009efficient,li2014reducing,yuan2014lightlda,yu2015scalable} exist to speed this up, as we shall see in Sec.~\ref{sec:analysis}.

\subsection{Sample from 
Probability Distributions}\label{sec:general-alg}
Before introducing the algorithm, we first describe some useful tools to sample from probability distributions, which are used by WarpLDA and other algorithms~\cite{yuan2014lightlda,li2014reducing}.

\textbf{Metropolis-Hastings (MH):}
Let $p(x)$ be an (unnormalized) target distribution. We consider the nontrivial case that it is hard to directly draw samples from $p(x)$. MH methods construct a Markov chain with an easy-to-sample \emph{proposal distribution} $q(\hat x_t | x_{t-1})$ at each step $t$. Starting with an arbitrary state $x_0$, MH repeatedly generates samples from the proposal distribution $\hat x_{t} \sim q(\hat x_{t} | x_{t-1})$, and updates the current state with the new sample with an \emph{acceptance rate} $\pi_t = \min\{1, \frac{p(\hat x_t) q(x_{t-1} | \hat x_t)}{p(x_{t-1}) q(\hat x_t | x_{t-1})}\}$. Under some mild conditions, it is guaranteed that $p(x_{t})$ converges to $p(x)$ as $t\rightarrow \infty$, regardless of $x_0$ and $q(\hat x | x)$ (See Alg.~\ref{alg:mh}). In LDA, $p(x)$ is the distribution of topic assignment in Eq.~(\ref{eqn:cgs}), whose sampling complexity is $O(K)$, and $q(\hat x_{t} | x_{t-1})$ is often a cheap approximation of $p(x)$, as will be clear soon.

\begin{algorithm}[t]
\caption{Metropolis-Hastings algorithm\label{alg:mh}}
\begin{algorithmic}
\Require Initial state $x_0$, $p(x)$, $q(\hat x | x)$, number of steps $M$
\For {$t\leftarrow 1 \mbox{ to } M$}
\State Draw $\hat x \sim q(\hat x | x_{t-1})$
\State Compute the acceptance rate $\pi = \min\{1, \frac{p(\hat x) q(x_{t-1} | \hat x)}{p(x_{t-1}) q(\hat x | x_{t-1})}\}$
\State $x_t = \begin{cases}
\hat x & \mbox{with probability $\pi$}\\
x_{t-1} & \mbox{otherwise}
\end{cases}$
\EndFor
\end{algorithmic}
\end{algorithm}

\textbf{Mixture of multinomials:} If a multinomial distribution has the form\vspace{-1em}
$$p(x=k) \propto A_k + B_k,$$
it can be represented by a mixture of two distributions, 
$$p(x=k) = \frac{Z_A}{Z_A+Z_B}p_A(x=k) + \frac{Z_B}{Z_A+Z_B}p_B(x=k),$$
where $Z_A=\sum_k A_k$ and $Z_B=\sum_k B_k$ are the normalizing coefficients, and $p_A(x=k)=\frac{A_k}{Z_A}, p_B(x=k)=\frac{B_k}{Z_B}$ are the normalized mixture distributions. By introducing an extra binary variable $u$, which follows a Bernoulli distribution $\mbox{Bern}(\frac{Z_A}{Z_A+Z_B})$, and defining $p(x | u=1) = p_A(x), p(x | u=0) = p_B(x)$, one can confirm that $p(x)$ is a marginal distribution of $p(u) p(x|u)$. Therefore, a sample from $p(x)$ can be drawn via an ancestral sampler, which first draws $u \sim \mbox{Bern}(\frac{Z_A}{Z_A+Z_B})$ and then samples $x$ from $p_A(x)$ if $u=1$ and from $p_B(x)$ if $u=0$. 
This principle is useful when both  $p_A(x)$ and $p_B(x)$ are easy to sample from.


\textbf{Alias Sampling:} Alias sampling~\cite{walker1977efficient} is a technique to draw samples from a $K$-dim multinomial distribution $p(x)$ in $O(1)$ after $O(K)$ construction of an auxiliary structure called alias table. The alias table has $K$ bins with equal probability, with at most two outcomes in each bin. 
The samples can be obtained by randomly selecting a bin and then randomly selecting an outcome from that bin. Alias sampling has been used in previous LDA works~\cite{li2014reducing,yuan2014lightlda}.


\section{Analysis of existing algorithms}\label{sec:analysis}
We now describe our methodology to analyze the memory access efficiency, and analyze of existing fast sampling algorithms of LDA~\cite{yao2009efficient, li2014scaling, yuan2014lightlda,yu2015scalable} to motivate the development of WarpLDA.

\subsection{Methodology}
The running time of LDA is often dominated by random memory accesses, whose efficiency depends on the average \emph{latency} of each access. As accessing main memory is very slow, modern computers exploit \emph{locality} to reduce the latency. If many memory accesses are concentrated in a small area which fits in the cache, the latency will reduce greatly, as shown in Table~\ref{tbl:cache-latency}. 
We focus on the L3 cache in this paper, which is about 30 megabytes, and is at least six times faster than the main memory.

The size of randomly accessed memory is an important factor to the latency, because a smaller memory region can fit in a higher-level cache, which has smaller latency. So we analyze the memory access efficiency by the size of randomly accessed memory, more precisely, \emph{the size of possibly accessed memory region when sampling the topic assignments for a document $\vect z_{d}$ or a word $\vect z_w$}, 
depending on whether the sampling algorithm visits the tokens \emph{document-by-document} or \emph{word-by-word}, which will be defined soon.


As we shall see soon, there are two main types of random access encountered for LDA: (1) randomly accessing a matrix of size $O(KV)$ or $O(DK)$; and (2) randomly accessing a vector of size $O(K)$. As $K$, $V$ and $D$ can all exceed one million, the matrix is typically tens of gigabytes in size even when it is sparse; but the vector is megabytes or smaller in size, and fits in the L3 cache. 
Therefore, it is reasonable to assume that random accesses to matrices are not efficient, while random accesses to vectors are efficient. 
 
Existing fast sampling algorithms for LDA follow CGS that we introduced in Sec.~\ref{sec:basics-of-lda}, which iteratively visits every token of the corpus, and samples the topic assignments based on the count matrices $\vect C_d$ and $\vect C_w$. The tokens can be visited in any order. Two commonly used ordering are \emph{document-by-document} which firstly visits all tokens for document 1, and then visits all tokens for document 2, and so on; and \emph{word-by-word}  which firstly visits all tokens for word 1, and then visits all tokens for word 2, and so on. These orderings determine which one of the two count matrices $\vect C_d$ and $\vect C_w$ can be accessed efficiently, as we will see in Sec.~\ref{sec:analysis-sub}.

\subsection{Existing Fast Sampling Algorithm}
We now summarize existing fast sampling algorithms~\cite{yao2009efficient,li2014reducing,yuan2014lightlda,yu2015scalable}.
These algorithms can be categorized as being either sparsity-aware or MH-based, and they have different factorizations to the basic CGS sampling formula Eq.~(\ref{eqn:cgs}).  For clarity all the $^{\neg dn}$ superscripts of CGS based algorithms are omitted.

Sparsity-aware algorithms utilize the sparse structure of the count matrices $\vect C_d$ and $\vect C_w$. For example, SparseLDA~\cite{yao2009efficient} has the factorization $C_{wk}\frac{C_{dk}+\alpha_k}{C_k+\bar\beta}$ $+\beta\frac{C_{dk}}{C_k+\bar\beta}+\frac{\alpha_k\beta}{C_k+\bar\beta}$ and it enumerates all non-zero entries of $\vect c_w$ and $\vect c_d$ to calculate the normalizing constant of each term, which are $\sum_{k=1}^K C_{wk}\frac{C_{dk}+\alpha_k}{C_k+\bar\beta}$, $\sum_{k=1}^K \beta\frac{C_{dk}}{C_k+\bar\beta}$ and $\sum_{k=1}^K \frac{\alpha_k\beta}{C_k+\bar\beta}$ respectively. 
AliasLDA~\cite{li2014reducing} has the factorization $C_{dk}\frac{C_{wk}+\beta}{C_k+\bar\beta}+\alpha_k\frac{C_{wk}+\beta}{C_k+\bar\beta}$, where $C_{wk}$ and $C_k$ in the latter term are approximated with their stale versions. AliasLDA enumerates all non-zero entries of $\vect c_d$ to calculate the normalizing constant of the first term,
and an alias table is used to draw samples from the latter term in amortized $O(1)$ time. Then, an MH step is used to correct the bias of stale topic counts. 
F+LDA~\cite{yu2015scalable} has the same factorization as AliasLDA but visits the tokens word-by-word, and use a F+ tree for the exact sampling of the latter term.  

MH-based algorithms rely on some easy-to-sample proposal distributions to explore the state space $\{1, \dots, K\}$. The proposal distributions are not necessarily sparse, and hence MH-based algorithms can be applied to models whose $p(z_{dn})$'s do not have sparsity structures, e.g., MedLDA~\cite{zheng2015linear} or dynamic topic models~\cite{bhadury2016scaling}.
For example, LightLDA~\cite{yuan2014lightlda} alternatively draws samples from two simple proposal distributions $q^{\mbox{doc}}\propto (C_{dk}+\alpha_k)$ and $q^{\mbox{word}}\propto \frac{C_{wk}+\beta}{C_k+\bar\beta}$, and accepts the  proposals by the corresponding acceptance rate. 
The time complexity of sampling for a token is $O(1)$. 
The $O(1)$ complexity of LightLDA has already reached the theoretical lower bound, however its practical throughput is only roughly 4M tokens per second per machine~\cite{yuan2014lightlda}, due to its slow random access.

\subsection{Analysis}\label{sec:analysis-sub}
All the aforementioned algorithms are not memory efficient because they have random accesses to large matrices. 
Specifically, the main random accesses of these algorithms are to the count matrices $\vect C_d$ and $\vect C_w$. Ignoring all details of computing, we only focus on the reading and writing to the count matrices $\vect C_d$ and $\vect C_w$. When sampling $z_{dn}$ for each token $w_{dn}$, the memory access pattern to the count matrices can be summarized as follows:
\begin{itemize}[label={}]
  \setlength\itemsep{0em}
    \item read $C_{dk}$ and $C_{wk}$, for $k\in \mathcal{K}_{dn}$;
    \item write $C_{dz_{dn}}$ and $C_{w_{dn}z_{dn}}$,
\end{itemize}
where $k\in \mathcal{K}_{dn}$ is a set. 
The existing algorithms differ in two aspects:
(1) {\it Ordering of visiting tokens}: SparseLDA, AliasLDA and LightLDA visit tokens document-by-document, while F+LDA visits tokens word-by-word;
and (2) {\it Set $\mathcal{K}_{dn}$}: The set $\mathcal{K}_{dn}$ depends on the sparsity structure of the proposal distribution. For SparseLDA, $\mathcal{K}_{dn}$ is the set of non-zero topics of $\vect c_w$ and $\vect c_d$; for AliasLDA and F+LDA, $\mathcal{K}_{dn}$ is the set of non-zero topics of $\vect c_d$; and for LightLDA, $\mathcal{K}_{dn}$ is the set of some samples from the proposal distribution.

We have two important observations, where the accesses can be made efficient if only $\vect C_d$ or $\vect C_w$ is accessed, 
while the accesses are inefficient if both $\vect C_d$ and $\vect C_w$ are accessed, as detailed below:

\textbf{Accessing either $\vect C_d$ or $\vect C_w$: } Without loss of generality, assume only $\vect C_d$ is accessed. There will be a lot of accesses to random entries of the matrix $\vect C_d$. However, these random accesses are sorted by $d$, if the tokens are visited document-by-document. For sampling the tokens in document $d$, only one row $\vect c_d$ is accessed. Therefore, the size of randomly accessed memory per-document is only $K_d$,  which is the size of $\vect c_d$, and fits in the L3 cache. Moreover, the rows except $\vect c_d$ need not to be stored in the memory and can be computed on-the-fly upon request, reducing the storage overhead. 
Symmetrically, if only $\vect C_w$ is accessed, the accesses can be restricted in a vector $\vect c_w$ by visiting the tokens word-by-word. The counts $\vect c_w$ can be computed on-the-fly as well.

\textbf{Accessing both $\vect C_d$ and $\vect C_w$: } Unfortunately, the accesses to $\vect C_d$ and $\vect C_w$ are coupled in all the aforementioned algorithms, i.e., they need to access both $\vect C_d$ and $\vect C_w$ when sampling a token. If the tokens are visited document-by-document, the accesses to $\vect C_w$ are not sorted by $w$, and spread in the large matrix $\vect C_w$ whose size is $O(KV)$.
If the tokens are visited word-by-word, the accesses to $\vect C_d$ are not sorted by $d$, and spread in the large matrix $\vect C_d$ whose size is $O(DK)$.
Thus, all the existing algorithms have $O(KV)$ or $O(DK)$ size of randomly accessed memory, which is not efficient. 

Table~\ref{tbl:summary-of-algorithms} summarizes the existing algorithms. Note that the amount of sequential accesses is smaller than or equals to the amount of random accesses for all the fast sampling algorithms.
This justifies our previous argument that the running time of LDA is dominated by random accesses based on the well-known fact that sequential accesses are faster than random accesses.

WarpLDA addresses the inefficient memory access problem by decoupling the accesses to $\vect C_{d}$ and $\vect C_{w}$. Particularly, WarpLDA first visits all tokens but only accesses $\vect C_d$, and then visits all tokens again but only accesses $\vect C_w$. After that, we can choose the corresponding ordering of visiting the tokens so that the accesses to both matrices can be restricted in the current row $\vect c_d$ and $\vect c_w$, without affecting the correctness of the algorithm.


\section{WarpLDA}

We now present WarpLDA, a novel MH-based algorithm that finds a maximum {\it a posteriori} (MAP) estimate of LDA with an $O(1)$ per-token sampling complexity and an $O(K)$ size of randomly accessed memory per-document (or word), by a designed reordering strategy to decouple the accesses to $\vect C_d$ and $\vect C_w$.


\subsection{Notations}

\begin{figure}[t]
\centering
	\includegraphics[width=\linewidth]{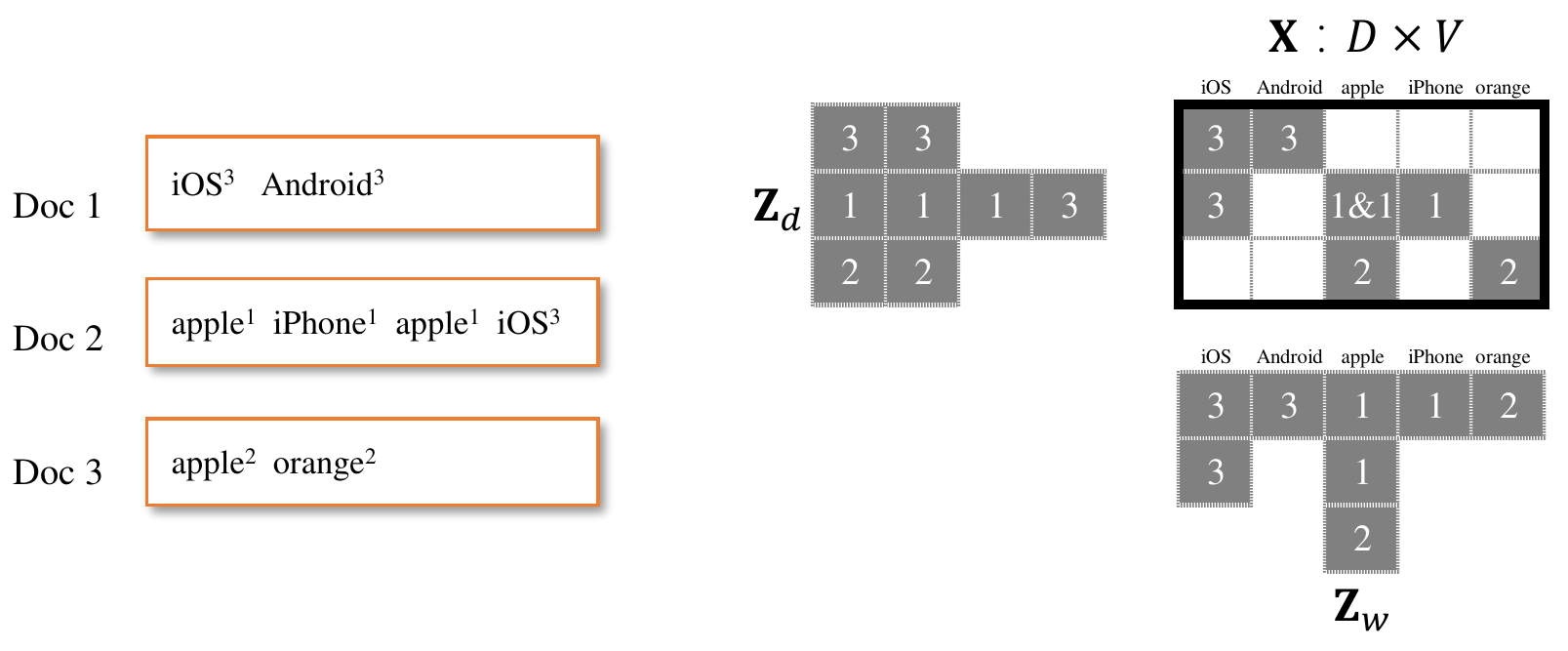}\vspace{-.3cm}
	\caption{Data representation of WarpLDA. Left: documents, the subscript of each token is its topic assignment $z_{dn}$. Right: the topic assignment matrix $\vect X$, $\vect Z_d$ and $\vect Z_w$. The \& sign separates multiple entries in one cell.
\label{fig:data-model}} \vspace{-.5cm}
\end{figure}

We first define some notations to make presentation clear.  
Specifically, we define a $D\times V$ \emph{topic assignment matrix} $\vect X$, where the topic assignment $z_{dn}$ is put in the cell $(d, w_{dn})$ (See Fig.~\ref{fig:data-model} for an illustration). Note that there might be multiple topic assignments in a single cell because a word can appear more than once in a document.
Let $\vect z_d$ be the collection of the $L_d$ topic assignments in the $d$-th row of $\vect X$, with $z_{dn}$ being its $n$-th element. The ordering of the  elements in $\vect z_d$ can be arbitrary because LDA is a bag-of-words model and ignores word ordering. This definition of $z_{dn}$ is consistent with that in Sec.~\ref{sec:basics-of-lda}, where $z_{dn}$ is the topic assignment of the $n$-th token of document $d$. Again, let $\vect Z = \vect Z_d = \{\vect z_d\}_{d=1}^D$. 
%
Similarly, let $\vect z_w$ be the collection of the topic assignments in the $w$-th column of $\vect X$, with size $L_w$ and $z_{wn}$ being its $n$-th element.  Note that $L_w$ is the \emph{term frequency} of word $w$, i.e., the number of times that $w$ appears in the entire corpus. Let $\vect Z_w = \{\vect z_w\}_{w=1}^V$.

Besides a topic assignment, WarpLDA has $M$ topic proposals for each token, denoted as $z_{dn}^{(i)}$ (or $z_{wn}^{(i)}$), where $i=1, \dots, M$. 
Using the above definition, we have $\vect z_d^{(i)}$ (or $\vect z_w^{(i)}$) for a document and $\vect Z_d^{(i)}$ (or $\vect Z_w^{(i)}$) for the whole corpus.

It is worth noting that most of these notations are for the ease of presentation, and in the actual implementation we only store $\vect Z_w, \vect Z_w^{(i)}$ and the global count vector $\vect c_k$ (Sec.~\ref{sec:system}). The other variables, including $\vect X, \vect Z_d, \vect C_d, \vect C_w$ are either \emph{views} of $(\vect Z_w, \vect Z_w^{(i)})$, or can be computed on-the-fly.

\subsection{MCEM Algorithm of LDA}

As analysed above, in the original CGS   Eq.~(\ref{eqn:cgs}) and existing fast algorithms, it is difficult to decouple the access to $\vect C_{d}$ and $\vect C_{w}$, because both counts need to be updated instantly after the sampling of every token. We develop our efficient WarpLDA based on a new Monte-Carlo Expectation Maximization (MCEM) algorithm which is similar with CGS, but both counts are fixed until the sampling of all tokens are finished. This scheme allows us to develop a reordering strategy to decouple the accesses to $\vect C_d$ and $\vect C_w$, and minimize the size of randomly accessed memory. 

Specifically, WarpLDA seeks an MAP solution of the latent variables $\vect \Theta$ and $\vect \Phi$, with the latent topic assignments $\vect Z$ integrated out: 
$$\hat {\vect\Theta}, \hat {\vect\Phi} = \argmax_{\vect\Theta, \vect\Phi} \log p(\vect\Theta, \vect\Phi | \vect W, \vect\alpha', \beta'),$$
where $\alphav^\prime$ and $\beta^\prime$ are the Dirichlet hyper-parameters. Asuncion et al.~\cite{asuncion2009smoothing} have shown that this MAP solution is almost identical with the solution of CGS, with proper hyper-parameters. Our empirical results in Sec.~\ref{sec:sensitivity} also support this conclusion.

Computing $\log p(\vect\Theta, \vect\Phi | \vect W, \vect\alpha', \beta')$ directly is expensive because it needs to enumerate all the $K$ possible topic assignments for each token. We therefore optimize its lower bound as a surrogate.
Let $q(\vect Z)$ be a variational distribution. Then, by Jensen's inequality, we get the lower bound $\mathcal{J}(\vect \Theta, \vect \Phi, q(\vect Z))$: 
\begin{align}
\log p(\vect\Theta, \vect\Phi | \vect W, \vect\alpha', \beta')\ge &\mathbb{E}_q[\log p(\vect W, \vect Z | \vect \Theta, \vect \Phi) - \log q(\vect Z)]\nonumber
\\&+ \log p(\vect \Theta | \vect \alpha') + \log p(\vect \Phi | \beta') \nonumber \\
\triangleq  & \mathcal{J}(\vect \Theta, \vect \Phi, q(\vect Z)). \label{eqn:Jensen}
\end{align}

We then develop an Expectation Maximization (EM) algorithm~\cite{dempster1977maximum} to find a local maximum of the posterior $p(\vect \Theta, \vect \Phi |\vect W, \vect \alpha^{\prime}, \beta^{\prime})$, where the E-step maximizes  $\mathcal{J}$ with respect to the variational distribution $q(\vect Z)$ and the M-step  maximizes $\mathcal{J}$ with respect the to model parameters $(\vect \Theta, \vect \Phi)$, while keeping $q(\vect Z)$ fixed. 
One can prove that the optimal solution at E-step is $q(\vect Z) = p(\vect Z | \vect W, \vect \Theta, \vect \Phi)$ without further assumption on $q$.
We apply Monte-Carlo approximation on the expectation in Eq.~(\ref{eqn:Jensen}),
\begin{align*}
\mathbb{E}_q[\log p(\vect W, \vect Z | \vect \Theta, \vect \Phi) - \log q(\vect Z)]\approx \\
\frac{1}{S}\sum_{s=1}^S\log p(\vect W, \vect Z^{(s)} | \vect \Theta, \vect \Phi) - \log q(\vect Z^{(s)}), \end{align*}
where $\vect Z^{(1)}, \dots, \vect Z^{(S)} \sim q(\vect Z)=p(\vect Z | \vect W, \vect \Theta, \vect \Phi).$ We set the sample size $S=1$ and use $\vect Z$ as an abbreviation of $\vect Z^{(1)}$.

 \noindent\textbf{Sampling $\vect Z$:} Each dimension of $\vect Z$ can be sampled independently,
\begin{align}
    q(z_{dn}=k)\propto p(\vect W, \vect Z | \vect \Theta, \vect \Phi) \propto \theta_{dk} \phi_{w_{dn}, k}\label{eqn:zori}.
\end{align}

\noindent\textbf{Optimizing $\vect \Theta, \vect \Phi$:} With the Monte-Carlo approximation,
\begin{align*}
    \mathcal{J} &\approx \log p(\vect W, \vect Z | \vect \Theta, \vect \Phi) + \log p(\vect \Theta | \vect \alpha^\prime) + \log p(\vect \Phi |  \beta^\prime) + \mbox{const.} \\
                &= \sum_{d, k} (C_{dk}+\alpha^\prime_k-1)\log \theta_{dk} \\
                &+ \sum_{k, w}(C_{kw}+\beta^\prime-1)\log \phi_{kw} + \mbox{const.},
\end{align*}
with the optimal solutions
\begin{align}
\hat{\theta}_{dk} \propto C_{dk} + \alpha^\prime_k - 1, ~~~~ 
\hat{\phi}_{wk} = \frac{C_{wk} + \beta^\prime - 1}{C_k + \bar\beta^\prime - V}.\label{eqn:phi-hat}
\end{align}

Instead of computing and storing $\hat{\vect \Theta}$ and $\hat{\vect \Phi}$, we compute and store $\vect C_{d}$ and $\vect C_{w}$ to save memory because the latter are sparse. Plug Eq.~(\ref{eqn:phi-hat}) to Eq.~(\ref{eqn:zori}), and let $\vect \alpha=\vect \alpha^\prime-1, \beta=\beta^\prime-1$, we get the full MCEM algorithm, which iteratively performs the following two steps until a given iteration number is reached:
\begin{itemize}
    \item E-step: Sample $z_{dn}\sim q(z_{dn}=k)$, where
        \begin{align}
            q(z_{dn}=k)\propto (C_{dk} + \alpha_k)\frac{C_{wk} + \beta_w}{C_k + \bar\beta} . \label{eqn:mcem}
        \end{align}

    \item M-step: Compute $\vect C_{d}$ and $\vect C_{w}$ by $\vect Z$.
\end{itemize}
Note the resemblance between Eq.~(\ref{eqn:mcem}) and Eq.~(\ref{eqn:cgs}) intuitively justifies why MCEM leads to similar results with CGS. The difference between MCEM and CGS is that MCEM updates the counts $\vect C_d$ and $\vect C_w$ after sampling all $z_{dn}$s, while CGS updates the counts instantly after sampling each $z_{dn}$.
The strategy that MCEM updates the counts after sampling all $z_{dn}$s is called \emph{delayed count update}, or simply \emph{delayed update}.
MCEM can be viewed as a CGS with delayed update, which has been widely used in existing algorithms~\cite{newman2007distributed,ahmed2012scalable}. While previous work uses delayed update as a trick, we hereby present a theoretical guarantee to converge to a MAP solution. Delayed update is important for us to 
decouple the accesses of $\vect C_d$ and $\vect C_w$
to improve cache locality, without affecting the correctness, as will be explained in Sec.~\ref{sec:org-the-comp}.

\subsection{Sampling the Topic Assignment}
A naive application of Eq.~(\ref{eqn:mcem}) is $O(K)$ per-token. We now develop an MH algorithm for faster sampling. 

Specifically, starting from an initial state $z_{dn}^{(0)}$, we draw samples alternatively from one of the two proposal distributions:
\begin{align}
q^{\mbox{doc}}(z_{dn}=k) &\propto C_{dk}+\alpha_k \nonumber \\
    q^{\mbox{word}}(z_{dn}=k) &\propto C_{wk}+\beta,\label{eq:WarpLDA-proposal}
\end{align}
and update the current state with the acceptance rates:
\begin{align}
\pi^{\mbox{doc}}_{k\rightarrow k'} &= \min \left\{1, \frac{C_{wk'} + \beta}{C_{wk} + \beta} \frac{C_k + \bar \beta}{C_{k'} + \bar \beta} \right\} \nonumber \\ 
\pi^{\mbox{word}}_{k\rightarrow k'} &= \min \left\{1, \frac{C_{dk'} + \alpha_{k'}}{C_{dk} + \alpha_k}\frac{C_k+ \bar \beta}{C_{k'}+ \bar \beta} \right\}.\label{eqn:warplda-accept-rate}
\end{align}
Similar as in Yuan et al.'s work~\cite{yuan2014lightlda}, we can prove that this scheme converges to the correct stationary distribution in Eq.~(\ref{eqn:mcem}). 

Computing the acceptance rates only involves constant number of arithmetic operations, so it is $O(1)$. 
The proposal distributions in  Eq.~(\ref{eq:WarpLDA-proposal}) are mixture of multinomials mentioned in Section~\ref{sec:general-alg}, with the mixing coefficient $\frac{L_d}{L_d+\bar\alpha}$.\footnote{Due to the symmetry we only consider $q^{\mbox{doc}}$.} There are two possible methods to draw samples from $p_A(z_{dn}=k)\propto C_{dk}$ in $O(1)$ time:
(1) {\bf Alias sampling}: 
Build a $K_d$-dim alias table for all the non-zero entries of $\vect c_d$;
and (2) {\bf Random positioning}: Noticing that $\vect c_d$ is the count of $\vect z_d$, randomly select a position $u\in \{1, \dots, L_d\}$, and return $z_{du}$. Alias sampling is also used to draw samples from $p_B(z_{dn}=k)\propto a_k$ in $O(1)$.
Because both sampling from proposal distributions and computing acceptance rates can be done in $O(1)$, the algorithm is $O(1)$ per-token.

\subsection{Reordering the Computation}\label{sec:org-the-comp}
As discussed above, the size of randomly accessed memory per-document or word is an important factor that influences the efficiency, but it is difficult to reduce for existing algorithms due to the coupling of the counts $\vect C_w$ and $\vect C_d$. Thanks to the delayed update strategy in MCEM, we are able to decouple the access to $\vect C_w$ and $\vect C_d$, and minimize the size of randomly accessed memory via a reordering strategy. Below, we explain how to do the reordering in the E-step and the M-step in turn.


\textbf{E-step:}
The E-step samples the topic assignment $z_{dn}$ for each token, while keeping the counts $\vect C_d, \vect C_w$ and $\vect c_k$ fixed.
Consider the sampling of a single topic assignment $z_{dn}$ with an MH algorithm. For simplicity, we only consider the document proposal $q^{\mbox{doc}}$. According to the MH Alg.~\ref{alg:mh}, starting with the initial state $z_{dn}$, we do the following in the $i$-th step ($i=1, \dots, M$):
\begin{itemize}\vspace{-.23cm}
    \item Draw the topic proposals $z_{dn}^{(i)}$ according to Eq.~(\ref{eq:WarpLDA-proposal}), where $q^{\mbox{doc}}(z_{dn}^{(i)}=k)\propto C_{dk}+\alpha_k$;\vspace{-.23cm}
    \item Update the current state $z_{dn}$ by the proposal $z_{dn}^{(i)}$, with the probability $\pi$, where $\pi = \min\{1, \frac{C_{wk^\prime}+\beta}{C_{wk}+\beta}\frac{C_k+\bar\beta}{C_{k^\prime}+\bar\beta}\}$, according to Eq.~(\ref{eqn:warplda-accept-rate}).\vspace{-.23cm}
\end{itemize}
Both $\vect C_d$ and $\vect C_w$ need to be accessed to sample $z_{dn}$ in the above procedure. Following our analysis in Sec.~\ref{sec:analysis-sub}, there are inevitable random accesses to matrices no matter whether we visit the tokens document-by-document or word-by-word. Thanks to the delayed update strategy in MCEM, we can address this problem by a reordering strategy.

Delayed update makes $\vect C_d$, $\vect C_w$ and $\vect c_k$ fixed during the entire E-step. An important corollary is that the proposal distribution $q^{\mbox{doc}}$, which depends solely on $\vect C_d$ and $\vect \alpha$, is fixed during the E-step. Therefore, we can draw the topic  proposals $z_{dn}^{(i)}$ at any time within the E-step, without affecting the correctness.
Particularly, we choose to draw the proposals for all tokens \emph{before} computing any acceptance rate. With this particular ordering, the sampling of all the topic assignments can be done in two separate steps:
\begin{enumerate}\vspace{-.23cm}
    \item Draw the topic proposals $z_{dn}^{(i)}$ for all tokens. This only accesses $\vect C_d$ and $\vect \alpha$.\vspace{-.23cm}
    \item Compute the acceptance rates and update the topic assignments for all tokens. This only accesses $\vect C_w$, $\vect c_k$ and $\beta$.\vspace{-.23cm}
\end{enumerate}
Since each step only accesses $\vect C_d$ or $\vect C_w$, following the analysis in Sec.~\ref{sec:analysis-sub}, we can make both steps memory efficient by carefully choosing the ordering of visiting tokens.  In the first step, tokens are visited document-by-document. Therefore,
when processing the $d$-th document, only a small vector $\vect c_d$ for the current document  is randomly accessed.  In the second step, the tokens are visited word-by-word, and when processing the $w$-th word, only a small vector $\vect c_w$ is randomly accessed. Because the vectors are small enough to fit in the L3 cache, WarpLDA is memory efficient.

\textbf{Word proposal:} The word proposal $q^{\mbox{word}}$ can be treated similarly as the doc proposal $q^{\mbox{doc}}$, by drawing the topic proposals for all tokens before computing any acceptance rate. There are also two separate steps for the sampling of all topic assignments: (1) Draw the topic proposals $z_{wn}^{(i)}$ for all tokens  (accesses $\vect C_w$ and $\beta$); and (2) Compute the acceptance rates and update the topic assignments for all tokens (accesses $\vect C_d$, $\vect c_k$, $\vect \alpha$ and $\beta$). 
These steps can be done efficiently by doing the first step word-by-word and doing the second step document-by-document. 

An WarpLDA iteration first samples the topic assignments for each token using the document proposal, and then samples again using the word proposal, which involves four passes of the tokens (two passes for each proposal). To improve the efficiency, these four passes can be compressed to two passes. The \emph{document phase} visits tokens document-by-document, and do the operations that require $\vect c_d$, which are computing the acceptance rates for the word proposal followed by drawing samples from the document proposal. Symmetrically, the \emph{word phase} visits tokens word-by-word, computes the acceptance rates for the document proposal, and then draws samples from the word proposal.

\textbf{M-step:}
Up to now, we  have talked about how to do the sampling of the topic assignments, i.e., the E-step of the MCEM algorithm. The M-step, which updates the counts $\vect C_d$, $\vect C_w$ and $\vect c_k$, need not to be conducted explicitly, because the counts can be computed on-the-fly. The only usage of the vector $\vect c_d$ is when processing document $d$ in the document phase. Hence, it can be computed by $\vect z_d$ when processing document $d$, and discarded after the document is finished. Similarly, the only usage of the row vector $\vect c_w$ is when processing word $w$ in the word phase, and it can be computed on-the-fly as well. Noticing $\vect c_k = \sum_d \vect c_d = \sum_w \vect c_w$, the count vector $\vect c_k$ can be accumulated when computing $\vect c_d$ for each document. 

The above facts also imply that we even need not to \emph{store} $\vect C_w$ and $\vect C_d$, which simplifies the system design as we shall see in Sec.~\ref{sec:system}, and again, justifies there are no random accesses to matrices --- we do not even store any of the matrices $\vect C_d$ and $\vect C_w$. 

\section{System Implementation}\label{sec:system}
WarpLDA not only improves the cache locality but also simplifies the distributed system design for training LDA on hundreds of machines. In previous systems for distributed LDA, including  parameter servers (PS)~\cite{ahmed2012scalable, li2014scaling} and model parallel systems~\cite{wang2014peacock,yuan2014lightlda},  all workers collaboratively refine a globally shared count matrix $\vect C_w$. This  adds additional complications to the system, such as read/write locks or delta threads.
WarpLDA is arguably simpler to implement because its only globally shared object is a small vector $\vect c_k$ which can be  updated and broadcast-ed to each machine in every iteration, and all the other data are local so that they can be processed independently by each worker.

In this section, we discuss the design of the distributed sparse matrix framework for WarpLDA. We then present techniques to implement the framework in a memory efficient fashion. Some application-level optimizations are also discussed.

\subsection{Programming Model}\label{sec:smce}

\begin{figure}
\begin{lstlisting}[language={[ANSI]C}]
template <class Data>
interface SparseMatrix {
    // Initialization
    void AddEntry(int r, int c, Data data);
    // Computation
    void VisitByRow(Operation op);
    void VisitByColumn(Operation op);
    // User-defined function
    interface Operation {
        operator () (vector<Data>& data);
    };
};
\end{lstlisting}\vspace{-.3cm}
\caption{Interface of a distributed sparse matrix. \label{fig:interface}}\vspace{-.3cm}
\end{figure}
We start the presentation of our framework for WarpLDA by abstracting its data structure.
In WarpLDA, the only data to manipulate are: 1) local per-token data, where each token $w_{dn}$ is associated with $M+1$ integers, which are the topic assignment and the topic proposals $y_{dn} \triangleq (z_{dn}, z_{dn}^{(1)}, \dots, z_{dn}^{(M)})$; and 2) global topic count vector $\vect c_k$. Noticeably, the count matrix $\vect C_w$ is computed on-the-fly and not stored, as we have analyzed in Sec.~\ref{sec:org-the-comp}. 

The local per-token data are stored in a $D\times V$ matrix $\vect Y$, where each token $w_{dn}$ corresponds to an entry at the position $(d, w_{dn})$ with $y_{dn}$ as its data. The matrix $\vect Y$ has the same structure with $\vect X$ in Fig.~\ref{fig:data-model}, but is augmented in the sense that each entry stores both a topic assignment and proposals. 

We now introduce the framework that we design for WarpLDA. The main data structure of our framework is the \emph{distributed sparse matrix} $\vect Y$ of size $D\times V$. There are some \emph{entries} in the sparse matrix, where each entry stores some \emph{data}. To help readers relate our presentation with the actual implementation, we use C++ notations for array indexing, e.g., \sys{\vect Y[d][w]} is the cell at the $d$-th row and $w$-th column of $\vect Y$. Let  \sys{\row[d]} as the collection of all entries in the $d$-th row, with  \sys{\row[d][n]} being its $n$-th element. Similarly, define the columns  \sys{\col[w]},   and \sys{\col[w][n]}. Let $a.size$ be the size of the collection $a$, such as $row[d].size$ and $col[d].size$.
There might be multiple entries in a same cell \sys{\vect Y[d][w]}. The structure of the matrix, i.e., the positions of the entries, is fixed, and only the data of the entries are iteratively refined to the solution. 

There are several methods to manipulate the matrix (See Fig.~\ref{fig:interface}): 
\begin{itemize}\vspace{-.25cm}
\item \texttt{AddEntry}: add an entry to a given cell \sys{\vect Y[d][w]} with the data \texttt{data}. This is called only at initialization time.\vspace{-.25cm}
\item \texttt{VisitByRow}: For each row $d$, update the data for each entry, given the data of all the entries in the current row \sys{row[d]}, i.e., \sys{\row[d]\leftarrow f_r(\row[d])}, where \sys{f_r(\cdot)} is a user defined function.\vspace{-.25cm}
\item \texttt{VisitByColumn}: For each column $w$, update the data for each entry, given the data of all the entries in the current column \sys{\col[w]}, i.e., \sys{\col[w]\leftarrow f_c(\col[w])}, where \sys{f_c(\cdot)} is another user defined function.\vspace{-.25cm}
\end{itemize}
Users initialize the matrix via the  \texttt{AddEntry} method, and then call \texttt{VisitByRow} and \texttt{VisitByColumn} with user defined functions for a number of times. The data of the matrix will be refined towards the result. 

For WarpLDA, the matrix $\vect Y$ stores the local per-token data $y_{dn}$, where each row is a document, and each column is a word. One can verify that the topic assignments of the $d$-th document $\vect z_d$ can be derived from the $d$-th row $row[d]$, and $row[d][n] = y_{dn} = (z_{dn}, z_{dn}^{(1)}, \dots, z_{dn}^{(M)})$. Similarly, $\vect z_w$ can be derived from $col[w]$.
To use the framework, we firstly initialize the matrix by adding an entry $y_{dn}$ to the cell $\vect Y[d][w_{dn}]$ for each token. The document phase can be  implemented with a single \texttt{VisitByRow}, where for each row (document), $\vect c_{d}$ is calculated with the topic assignments in $row[d]$; then the topic assignments in $row[d]$ are updated given the topic proposals and $\vect c_d$; and finally new proposals are created given new topic assignments. \footnote{Refer to  Alg.~\ref{alg:warplda} in the Appendix for details.} Similarly, the word phase can be implemented by \texttt{VisitByColumn}. The globally shared vector $\vect c_k=\sum_w \vect c_w$ can be updated by a simple ``reduce'' operation, which aggregates the word-topic count vectors $\vect c_w$, in the user defined function $f_c(\cdot)$. Since updating $\vect c_k$ can be implemented by the user, our framework does not handle the global count vector $\vect c_k$ but only $\vect Y$.

A basic implementation of this framework is MapReduce. Each entry is represented by a $(d, w)\rightarrow \vect Y[d][w]$ pair, and \texttt{VisitByRow} is implemented by two steps: 1) take the entries $(d, w)\rightarrow Y[d][w]$, aggregate them by row, and emit the rows $d\rightarrow row[d]$; 2) take the rows (documents) $d\rightarrow row[d]$, do the update $row[d]\leftarrow f_r(row[d])$ and emit the individual entries $(d, w)\rightarrow \vect Y[d][w]$. \texttt{VisitByColumn} is implemented by two similar steps, but aggregate the entries by column.
This implementation is useful for industrial users who want to build a simple distributed $O(1)$ LDA on top of the existing MapReduce framework. However, the shuffling overhead of MapReduce is often too large and the performance might be unsatisfactory. Hence, we present a dedicated implementation.




\subsection{Data Layout}
\label{sec:single-thread}
The first problem we need to address is how the matrix is represented internally. \texttt{VisitByRow} (or \texttt{VisitByColumn}) requires to provide $row[d]$ (or $col[w]$) to the user defined functions, while both $row[d]$ and $col[w]$ should be accessed efficiently, and the change of either should reflect to the underlying matrix.

There are a number of formats for a sparse matrix. Two most well-known examples are the Compressed Sparse Row (CSR) format and the Compressed Sparse Column (CSC) format. In the CSR representation $\vect Y_{CSR}$, the rows are stored continuously, and the accesses to each row is sequential. Similarly, in the CSC representation $\vect Y_{CSC}$, the accesses to each column is sequential.

One possible data layout is storing both $\vect Y_{CSR}$ and $\vect Y_{CSC}$. After each \texttt{VisitByRow} or \texttt{VisitByColumn} operation, perform a ``transpose'' operation to synchronize the change of one representation to the other representation. By storing both $\vect Y_{CSR}$ and $\vect Y_{CSC}$, the user defined operation always has sequential accesses to the data and hence is memory efficient. However, the transpose operation requires an extra pass of data which is expensive.

In WarpLDA, we avoid explicitly transposing  the matrix by only storing $\vect Y_{CSC}$. For  accessing the rows of the matrix, we store $\vect P_{CSR}$, which are \emph{pointers} to the entries in $\vect Y_{CSC}$. Since only one copy of the data is stored, there is no need for explicitly  transposing the matrix. However, the rows are accessed by indirect memory accesses, which are not sequential. We now show these indirect memory accesses are still memory efficient because the cache lines are fully utilized. 

For each column $col[w]$,  we sort the entries by their row id. Then, while in \texttt{VisitByRow}, the entries of a particular column $col[w]$ are always accessed sequentially, i.e., $col[w][i]$ is always accessed before $col[w][j]$ for all $i<j$. When an  entry $col[w][i]$ is accessed, the cache line containing it is fetched to the cache, which also contains the next few entries $(col[w][i+1], col[w][i+2], \dots)$. As long as the cache is large enough to store one cache line for each column, the cache lines can stay in the cache until all the entries on it are accessed. Thus, the cache lines are fully utilized. Moreover, the size of the columns follows the power-law, because they are term-frequencies of words in natural corpora~\cite{kingsley1932selective}. Therefore, the required cache size can be even smaller, comparable to the number of columns which have most entries of the sparse matrix. For example, in the ClueWeb12 corpus where the vocabulary size (number of columns) is 1,000,000, the first 10,000 words (columns) attributes to 80\% of the entries, and storing a cache line for these words (columns) requires only $10000\times 64\mbox{B}=625\mbox{KB}$.

\subsection{Scaling out}\label{sec:scaling-up}
WarpLDA can be scaled out to hundreds of machines to meet the requirements of learning large models on massive-scale corpora. We now present the key components for scaling out WarpLDA, including task partitioning, data placement and communication.

\subsubsection{Multi-threading and NUMA}
We address the data race and NUMA issues which are often encountered in multi threaded environments. 

WarpLDA is embarrassingly parallel because the workers operate on disjoint sets of data. To parallelize WarpLDA we only need to invoke the user defined functions in parallel for different rows and columns. 
In contrast, traditional frameworks such as Yahoo!LDA~\cite{ahmed2012scalable} and LightLDA~\cite{yuan2014lightlda} need to update the count matrix $\vect C_{w}$ in parallel, and require extra treatments such as read/write locks and delta threads. 

Modern computers have non-uniform memory access (NUMA), where each main memory DIMM belongs to a specific CPU socket. If one CPU needs to access data in the memory belongs to another CPU socket, the data flows through another CPU socket, resulting in degraded performance. For better performance, WarpLDA partitions the data $Y_{CSC}$ by column and the points $P_{CSR}$ by row, and bind each slice to a different CPU socket. Both the \texttt{VisitByColumn} and the visiting of the pointers in $P_{CSR}$ access only the local data, but the indirect memory accesses in \texttt{VisitByRow} may flow through other CPUs.


\subsubsection{Fine-grained Distributed Computing}
Distributed computation typically involves partitioning the task and addressing the communications between workers. We take a two-level partitioning strategy to overlap the computation and communication, and propose a greedy approach for balanced partitioning on the challenging case where the length of columns are distributed as power-law. The communications are implemented with \texttt{MPI\_Ialltoall} in Message Passing Interface (MPI).

\begin{figure}\vspace{-.5cm}
\centering
\includegraphics[width=\linewidth]{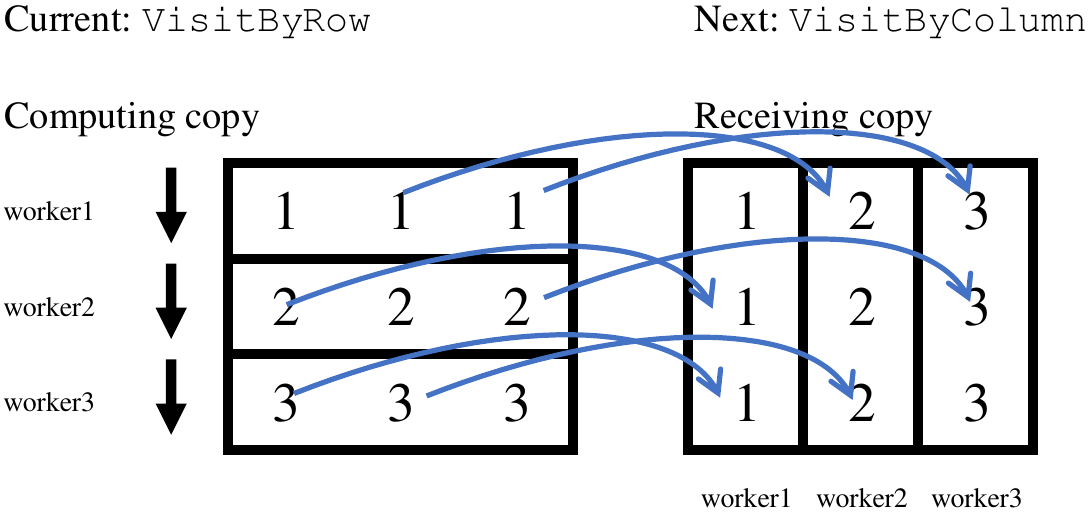}\vspace{-.3cm}
\caption{Data partitioning. The number indicates the worker each partition belongs to, the blue arrows are data flows during communication, and the black arrows are the  directions of visiting data.
\label{fig:mpi}}\vspace{-.3cm}
\end{figure}
Each \texttt{VisitByRow} and \texttt{VisitByColumn} involves a pass of the matrix $\vect Y$. To assign the task to different workers, we split the $D\times V$ data matrix $\vect Y$ as $P\times P$ \emph{partitions}, where $P$ is the number of MPI workers, and $P_{ij}$ is the $(i, j)$-th partition. 
In \texttt{VisitByRow}, the data is partitioned by row, i.e., worker $i$ has partitions $P_{i1}, \dots, P_{iP}$ in its memory; in \texttt{VisitByColumn}, the data is partitioned by column, i.e., worker $i$ has partitions $P_{1i}, \dots, P_{Pi}$ in its memory. Exchange of data happens only when the adjacent two operations are different, i.e., one is \texttt{VisitByRow} and the other is \texttt{VisitByColumn}, in which case the partitions are sent to their corresponding workers of the next operation after the current operation is finished. Because MPI requires both sending  buffer and receiving buffer, we maintain a computing copy and receiving copy of the data. For example, in Fig.~\ref{fig:mpi}, the current operation is \texttt{VisitByRow}, which can be conducted independently on each worker without any communications, due to our partitioning strategy. After the current  \texttt{VisitByRow} is finished, we need to partition the matrix by column for the next \texttt{VisitByColumn}, so we send partition $P_{ij}$ from worker $i$ to worker $j$, with \texttt{MPI\_Ialltoall}. If the next operation is \texttt{VisitByRow} instead, no communication is needed.

We can overlap the communication and computation by further divide each partition  as $B\times B$ \emph{blocks}, where $B \in [2, 10]$ is a small constant. 
During training, each of the blocks may be in one of the four states:
1) not started; 2) computing; 3) sending; 4) finished. In each \texttt{VisitByRow} / \texttt{VisitByColumn}, workers scan their blocks by row or column, and each block can be immediately sent to the receiving copy after it is finished.

\begin{figure}[t]
\centering
\includegraphics[width=0.7\linewidth]{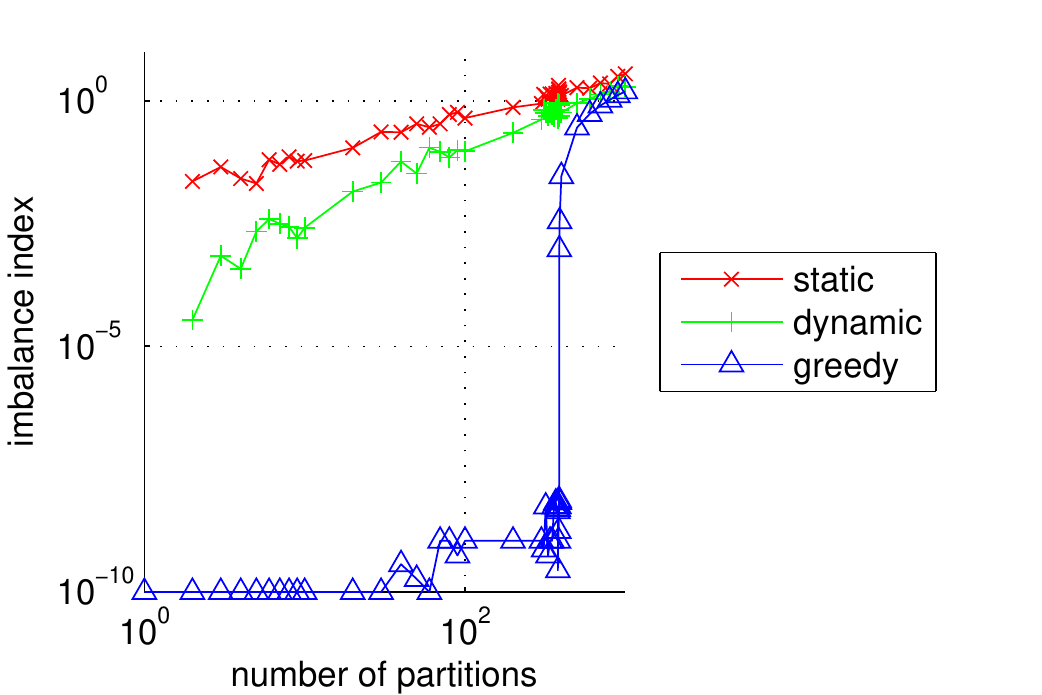}\vspace{-.3cm}
\caption{Comparison of partition strategies on the ClueWeb12 dataset. \label{fig:balance}}\vspace{-.3cm}
\end{figure}

To minimize the wait time, the number of tokens within each worker should be roughly the same, which implies that the sum of the number of tokens of all partitions within each row or column should be roughly the same. Note that the rows / columns can be treated independently: we partition the rows as $P$ disjoint sets $R_1, \dots, R_P$ and the columns as disjoint sets $C_1, \dots, C_P$, so that each set has roughly the same number of tokens. Given the partitioning of rows and columns, the $(i, j)$-th partition $P_{ij} = R_i\cap C_j$. 

Balanced partitioning of the columns  is challenging because the term frequencies of words from a natural corpus, i.e., $col[w].size$, can be very imbalanced
because of the power-law~\cite{kingsley1932selective}. For example, the most frequent word in the ClueWeb12 corpus occupies 0.257\% of all occurrences, after the removal of stop words. Considering each slice may only have 1\% of the tokens if there are 100 slices, this is a very large proportion, and random partitioning can be highly imbalanced in this case. We propose a greedy algorithm for balanced partitioning. First, all the words are sorted by their frequency in a decreasing order. Then from the most frequent word, we put each word (i.e., column) in the partition with the least number of total tokens. Because there are many low frequency words (the long tail), this algorithm can produce very balanced results. We compared the \emph{imbalance index} of our greedy algorithm with two randomized algorithms: \emph{static} first random shuffle the words, then partition so that each partition has equal number of words; \emph{dynamic} allows each partition to have different number of words, but each slice is continuous. Imbalance index is defined as $$\frac{\mbox{number of tokens in the largest partition}}{\mbox{average number of tokens of each partition}} - 1.$$
In the ideal case the imbalance index should be zero. Fig.~\ref{fig:balance} shows the experimental results on the ClueWeb12 corpus, where we can see that the greedy algorithm is much better than both randomized algorithms. The imbalance index of the greedy algorithm grows dramatically when the number of machines reach a few hundreds, because the size of the largest column is so large that it is impossible to  partition the columns in balance.


\subsection{Application Level Optimizations}
Besides system level optimizations, there are also some application level optimizations (i.e., optimization for the user defined function) for even better performance of WarpLDA. We describe them in this subsection. 

\textbf{Sparse vectors as hash tables: }
When $K$ is large, it is likely to have $K_d \ll K$ and $K_w \ll K$. It is more effective to use hash tables rather than dense arrays for the counts $\vect c_{d}$ and $\vect c_{w}$, because the size of a hash table is much smaller than that of a dense array, thus the cost of clearing the counts is smaller, and the size of randomly accessed memory is smaller. We choose an open addressing hash table with linear probing for hash collisions. The hash function is a simple \texttt{and} function, and the capacity is set to the minimum power of 2 that is larger than $\min\{K, 2L_d\}$ or $\min\{K, 2L_w\}$. We find that the hash table is almost as fast as a dense array even when $L_d > K$. Although LightLDA~\cite{yuan2014lightlda} also uses hash tables to store the counts, it is mainly for reducing the storage overhead instead of improving cache locality, because its size of randomly accessed memory is too large for the cache anyway.

\textbf{Intra-word parallelism: }
For the most frequent words in the corpus, $L_w$, i.e., the term frequency can be extremely large (e.g., tens of millions), so $L_w\gg K$. It is desirable to exploit \emph{intra-word parallelism} in this case.  Making all the threads working at the same column is beneficial for better cache locality, because only the $\vect c_w$ for \emph{one} word needs to be stored in the cache. Moreover, it helps balancing the load between workers in the case of $col[w].size$ is too large. We choose to exploit intra-word parallelism for words which $L_w > K$. As the user defined function is itself parallel in this case, we let the framework only invoke one user defined function at a time.

To parallelize the user defined function, we compute the count $\vect c_w$ on each thread, aggregate the results, and construct the alias table by a concurrent vector. Updating the topic assignments is embarrassingly parallel.


\section{Experiments}
\begin{table}[t]
\centering
\caption{Statistics of various datasets, where $T$ is the total number of words in the corpus.\label{tbl:datasets}}
\begin{tabular}{l | lllll}
\hline
Dataset & $D$ & $T$ & $V$  & $T/D$ \\ \hline
NYTimes & 300K & 100M & 102K  & 332 \\
PubMed & 8.2M & 738M & 141K  & 90 \\
\hline
ClueWeb12 (subset) & 38M & 14B & 1M & 367 \\
ClueWeb12 & 639M & 236B  & 1M & 378\\
\hline
\end{tabular}\vspace{-.3cm}
\end{table}

We now present empirical studies of WarpLDA, by comparing it with two strong baselines LightLDA~\cite{yuan2014lightlda} and F+LDA~\cite{yu2015scalable}. LightLDA is the fastest MH based algorithm, and F+LDA is the fastest sparsity-aware algorithm.
We compare with them in both time efficiency and the quality of convergence. 

\begin{figure*}[t]
\centering
\includegraphics[width=\linewidth]{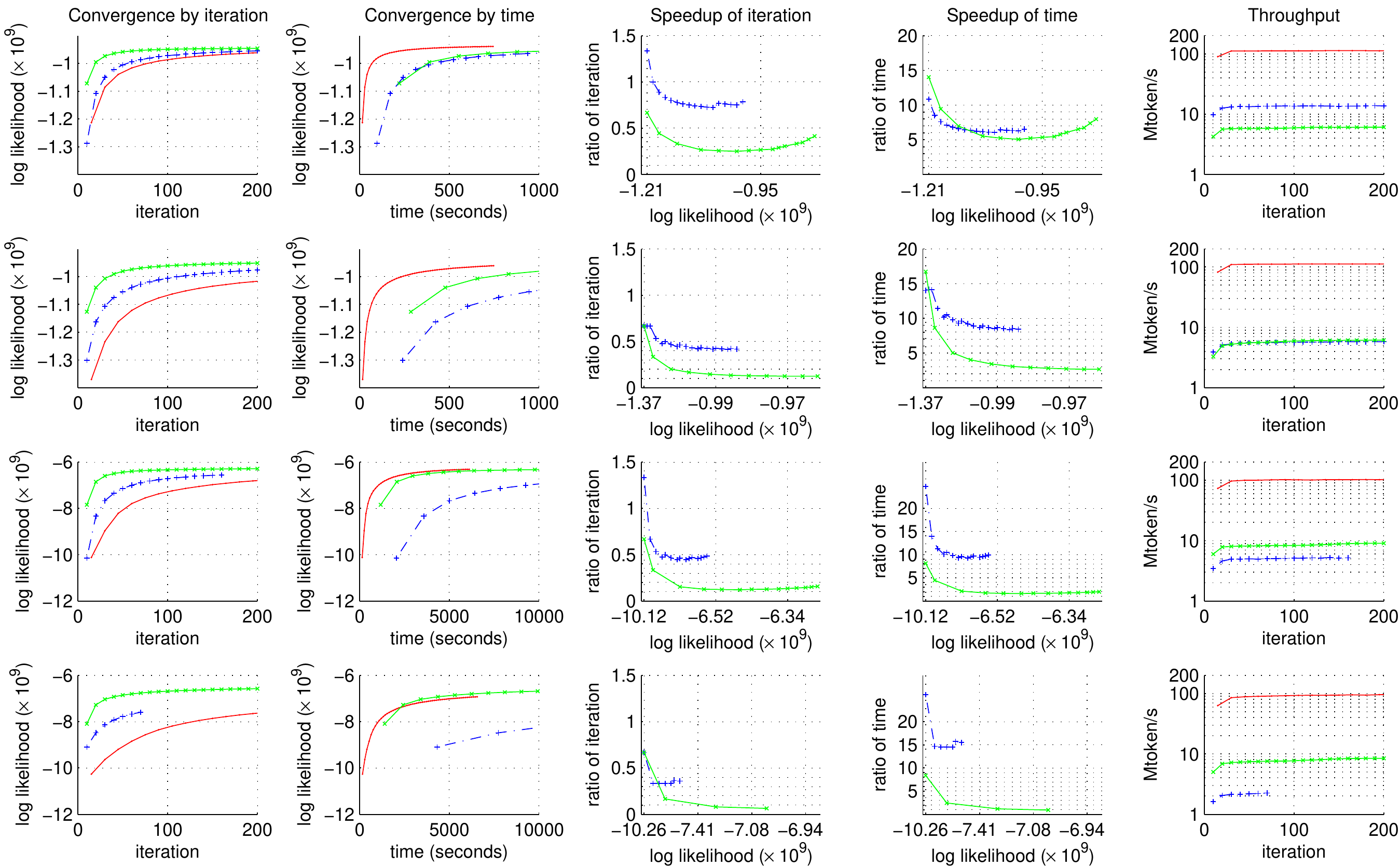}
\includegraphics[scale=.8]{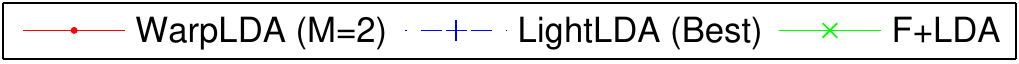}\vspace{-.3cm}
    \caption{Convergence results on NYTimes (1st row: $K=10^3$; 2nd row: $K=10^4$) and PubMed (3rd row: $K=10^4$; 4th row: $K=10^5$). Each column corresponds to an evaluation metric (Please see text for details). 
    The x-axis of column 3 and 4 are distorted for better resolution. 
    \label{fig:convergence}}\vspace{-.5cm}
\end{figure*}

\subsection{Datasets and Setups}
Table~\ref{tbl:datasets} summarizes the datasets. 
NYTimes and PubMed are standard datasets from the UCI machine learning repository~\cite{asuncion2007uci}; they consist of news articles and biomedical literature abstracts, respectively. ClueWeb12 is a large crawl of web pages.\footnote{\url{http://www.lemurproject.org/clueweb12.php/}} While NYTimes and PubMed are already tokenized, we extract text from ClueWeb12 using JSoup, remove everything except alphabets and digits, convert letters to lower case, tokenize the text by space and remove stop words. ClueWeb12 (subset) is a random subset of the full ClueWeb12.

The experiments are conducted on the Tianhe-2 supercomputer. Each node is equipped with two Xeon E5-2692v2 CPUs ($2\times 12$ 2.2GHz cores), and 64GB memory. Nodes are connected with InfiniBand, and single machine experiments are done with one node.

We set the hyper-parameters $\alpha=50/K$ and $\beta=0.01$. Following the previous work~\cite{ahmed2012scalable,yuan2014lightlda}, we measure the model quality by the widely adopted log joint likelihood (log likelihood in short):
\begin{align*}
\mathcal L = \log p(\vect W, \vect Z | \vect \alpha, \beta) 
=& 
\log \prod_d [\frac{\Gamma(\bar \alpha)}{\Gamma(\bar\alpha + L_d)} \prod_k \frac{\Gamma(\alpha_k+C_{dk})}{\Gamma(\alpha_k)}]\\
& \prod_k [ \frac{\Gamma(\bar\beta)}{\Gamma(\bar\beta+C_k)} \prod_w \frac{\Gamma(\beta+C_{kw})}{\Gamma(\beta)} ].
\end{align*}

\subsection{Speed of Convergence}\label{sec:convergence}
We first analyze the convergence behaviors. Fig.~\ref{fig:convergence} presents the single-machine results on the moderate-sized corpora, including NYTimes (first two rows) and PubMed (last two rows). We compare WarpLDA with a fixed $M=2$ to both LightLDA and F+LDA. Each algorithm is run for a fixed number of iterations. LightLDA is sensitive with $M$ because it affects the locality; We therefore pick up the $M$ that leads to fastest convergence over time for LightLDA, which is $M=4$ for the NYTimes dataset when $K=10^3$, $M=8$ for NYTimes when $K=10^4$, $M=8$ for PubMed when $K=10^4$, and $M=16$ for PubMed when $K=10^5$. A larger $M$ leads to longer running time per iteration, but the total number of iterations decreases in order to converge to a particular log-likelihood.

To have a full understanding, a diverse range of evaluation metrics are considered, including log-likelihood w.r.t the number of iterations ({\it 1st column}), log-likelihood w.r.t running time ({\it 2nd column}), the ratio of the iteration number of LightLDA (or F+LDA) over that of WarpLDA to get a particular log-likelihood ({\it 3rd column}), the ratio of running time of LightLDA (or F+LDA) over that of WarpLDA to get a particular log-likelihood ({\it 4th column}), and finally the throughput w.r.t the number of iterations ({\it 5th column}). From the results, we have the following observations: 

\begin{itemize}\vspace{-.23cm}
\item WarpLDA converges to the same log-likelihood as other baselines (1st and 2nd columns), demonstrating the good quality; \vspace{-.26cm}

\item WarpLDA converges faster than F+LDA and LightLDA in terms of running time (2nd column), despite that it needs more iterations than the competitorsm to reach the same likelihood (1st column). Overall, WarpLDA is consistently 5-15x faster than LightLDA in all evaluations, and is faster than F+LDA when $K\le 10^4$;\vspace{-.15cm}

\item Finally, WarpLDA is efficient --- WarpLDA achieves 110M token/s throughput on these datasets with a \emph{single machine}, much higher than that reported in previous works~\cite{ahmed2012scalable,yuan2014lightlda,yu2015scalable}, which is typically less than 10M token/s. \vspace{-.23cm}
\end{itemize}
Note that the convergence speed of F+LDA surpasses WarpLDA in later stages for the PubMed dataset when $K=10^5$. This is mainly due to the difference between sparsity aware algorithms and MH-based algorithms. As $K$ and $D$ are increasing, the vector $\vect c_w$ becomes less concentrated so that MH-based algorithms require more samples from $q^{\mbox{word}}$ to explore the state space, while the time complexity of the (exact) sparsity aware algorithms depends only on $K_d$ which is upper bounded by $L_d$. 

\begin{table}[]
\centering
\caption{L3 cache miss rate comparison, $M=1$. \label{tbl:cachemiss}}\vspace{-.3cm}
\begin{tabular}{l|c|c|c}
\hline
        Setting  &          LightLDA   & F+LDA     & WarpLDA     \\ \hline
NYTimes, $K=10^3$ & 33\% & 77\% & \textbf{17\%} \\ \hline
NYTimes, $K=10^4$ & 35\% & 53\% & \textbf{13\%} \\ \hline
PubMed, $K=10^4$ & 38\% & 57\% & \textbf{13\%} \\ \hline
PubMed, $K=10^5$ & 37\% & 17\% & \textbf{5\%} \\ \hline
\end{tabular}
\end{table}

\begin{figure*}[t]
\centering
\begin{minipage}[t]{0.31\linewidth}
\includegraphics[width=\linewidth]{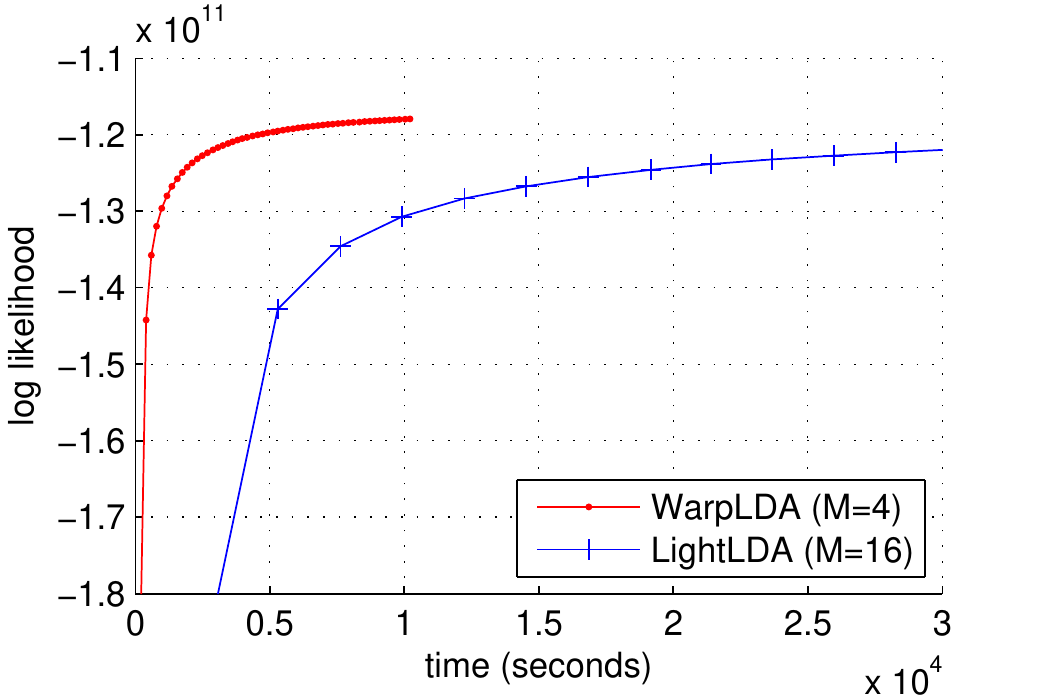}
\caption{Convergence on ClueWeb12 (subset), $K=10^4$. \label{fig:dist-convergence}}
\end{minipage}
\hfill
\begin{minipage}[t]{0.31\linewidth}
\includegraphics[width=\linewidth]{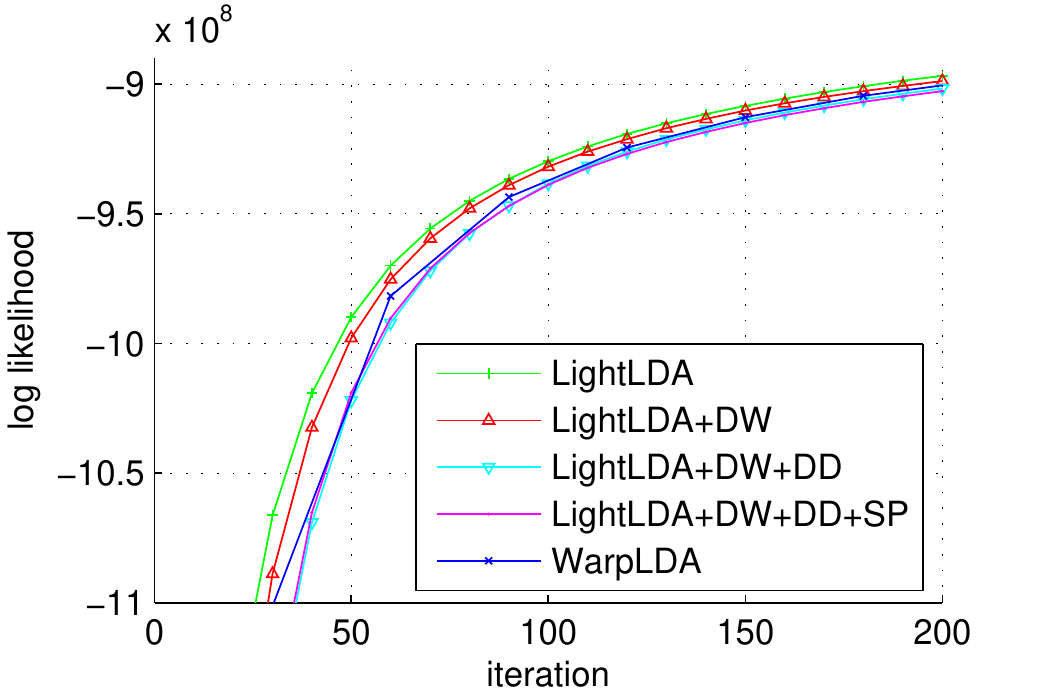}
\caption{Quality of the MCEM solution of WarpLDA and the CGS solution of LightLDA. NYTimes, $K=10^3$.\label{fig:impact-of-approximations}}
\end{minipage}
\hfill
\begin{minipage}[t]{0.31\linewidth}
\includegraphics[width=\linewidth]{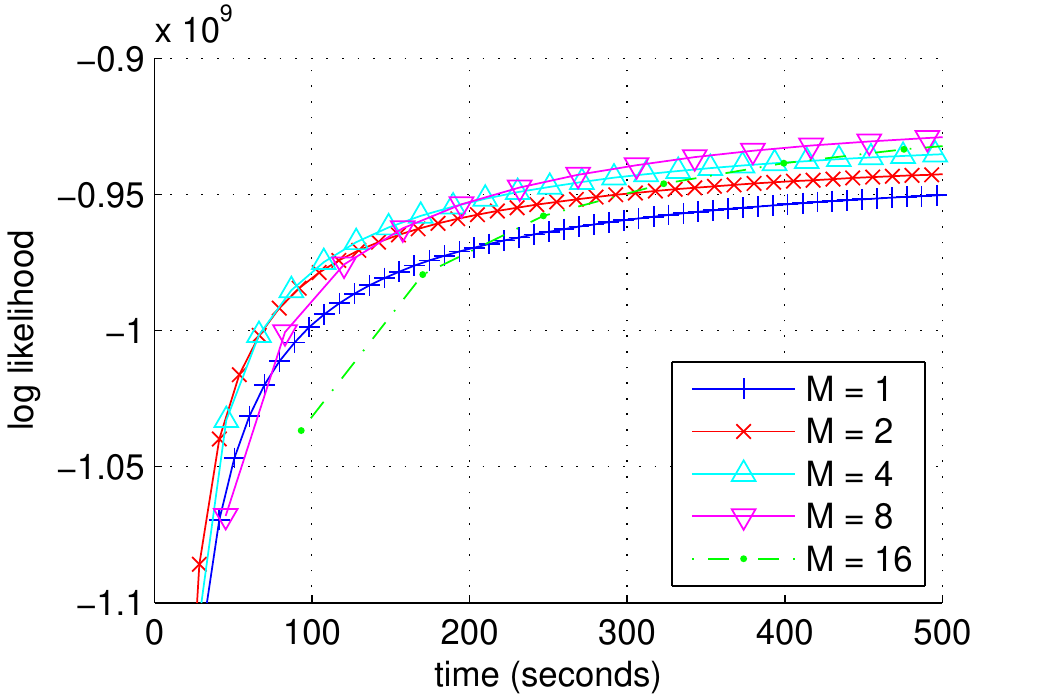}
\caption{Impact of different $M$. \label{fig:sensitivity}}
\end{minipage}
\end{figure*}

To show that WarpLDA is memory efficient, we compare the L3 cache miss rate of WarpLDA with that of LightLDA and F+LDA on the NYTimes and PubMed corpora. 
The cache miss rate is measured by PAPI. 
$M$ is set to 1 for both WarpLDA and LightLDA. From Table~\ref{tbl:cachemiss}, we can see that the L3 cache miss rate of WarpLDA is much lower than that of the competitors. This is  reasonable because the size of randomly accessed memory per-document for WarpLDA fits in the L3 cache, while the competitors require to randomly access a large matrix with tens-of-gigabytes in size.

For the distributed setting, WarpLDA ($M=4$) and LightLDA ($M=16$) are compared on a 38-million-document  ClueWeb12 (subset). Both algorithms are run on 32 machines. Fig.~\ref{fig:dist-convergence} presents the results. We can see that WarpLDA is about 10x faster than LightLDA to reach the same log-likelihood.


\begin{figure*}[t]
\centering
    \subfigure[Single machine]{\includegraphics[width=.24\linewidth]{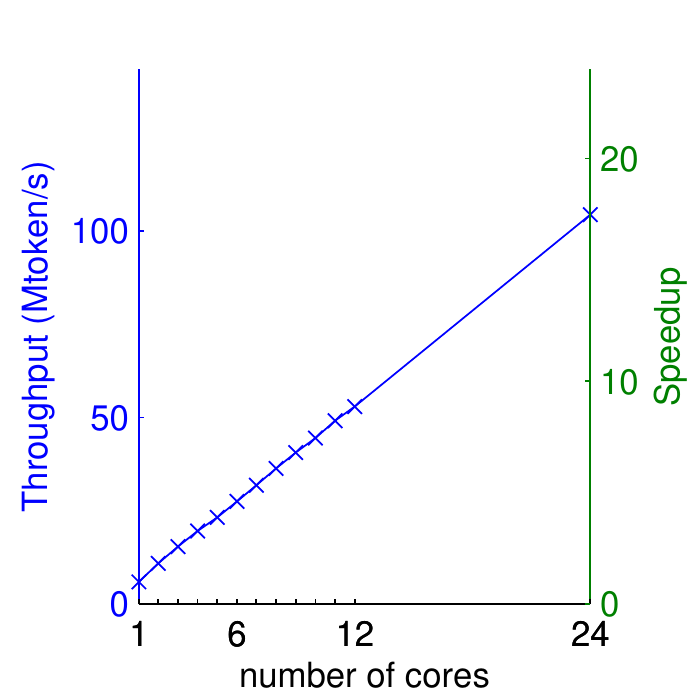}}
    \subfigure[Distributed]{\includegraphics[width=.24\linewidth]{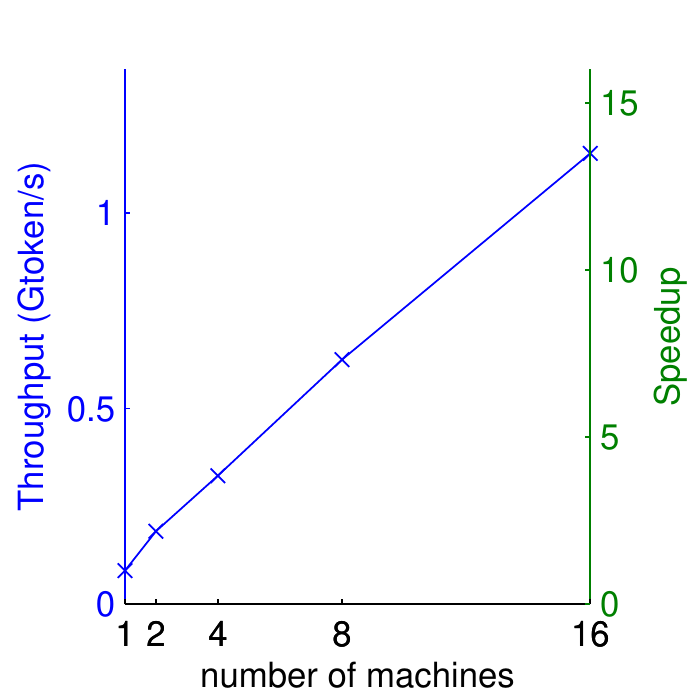}}
    \subfigure[Full ClueWeb12, convergence]{\includegraphics[width=0.24\linewidth]{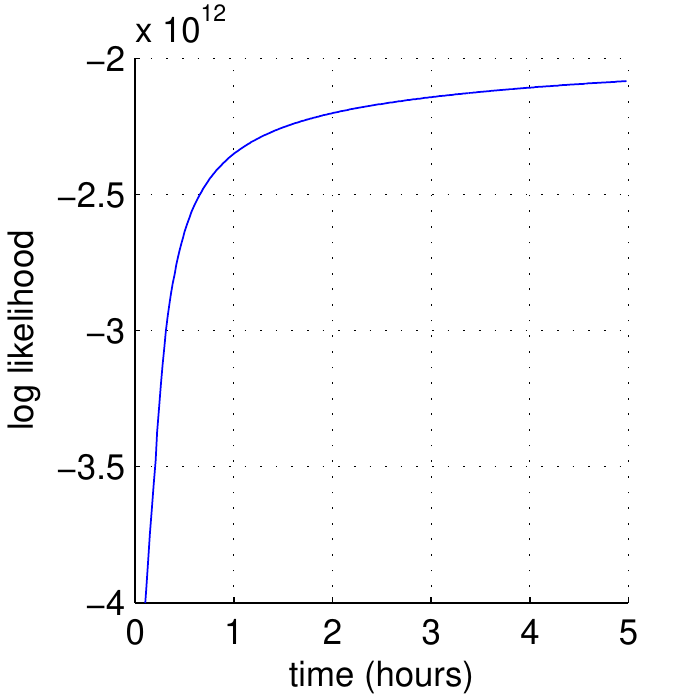}}
    \subfigure[Full ClueWeb12, throughput]{\includegraphics[width=0.24\linewidth]{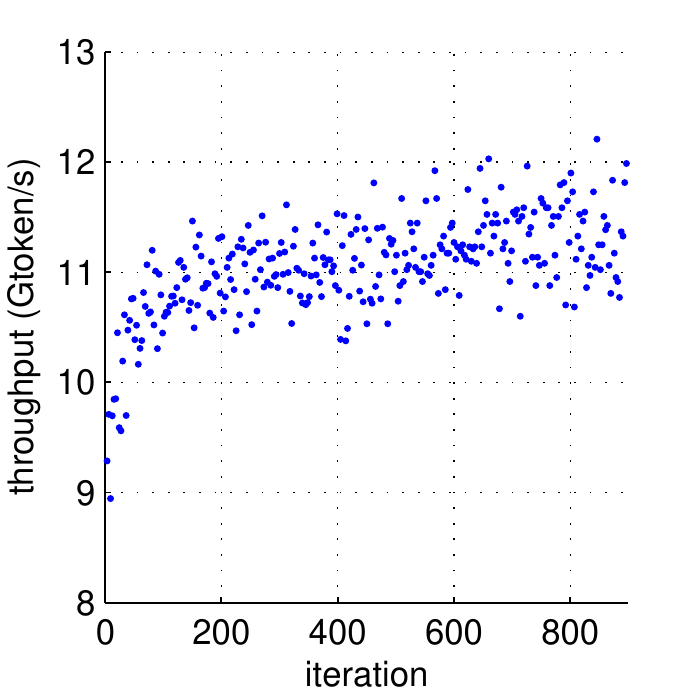}}\vspace{-.3cm}
	\caption{Scalability results. a) multi-threading speedup on NYTimes, $K=10^3$, $M=4$; 
	b) distributed speedup on PubMed, $K=10^4$, $M=1$;
	c, d) convergence and throughput on ClueWeb12, $K=10^6$, $MH=4$.\label{fig:scalability}}\vspace{-.3cm}
\end{figure*}

\subsection{Quality of Solutions}\label{sec:sensitivity}

Now we carefully analyze the quality of the solutions by WarpLDA, and show that the MCEM solution of WarpLDA is very similar with the CGS solution of LightLDA. Comparing the updates of LightLDA~\cite{yuan2014lightlda} and WarpLDA, we conclude that the differences that may affect the quality of the solutions are: (1) \emph{delayed count update}: the counts $\vect C_k$ and $\vect C_d$ are updated instantly in LightLDA but delayed in WarpLDA, (2) \emph{$q^{\mbox{word}}$}: For LightLDA $q^{\mbox{word}}\propto \frac{C_{wk}+\beta}{C_k +\bar\beta}$ and for WarpLDA $q^{\mbox{word}}\propto C_{wk}+\beta$.
Therefore, we vary these factors to gain insight on how each factor influences the convergence rate.
We use the NYTimes corpus as an example and set $K=10^3$ and $M=1$. The algorithms in comparison are
\begin{itemize}
    \item LightLDA: LightLDA with $M=1$, in which $\vect C_{d}$ is updated instantly, and $\vect C_{w}$ is updated every 300 documents (almost instantly). 
\item LightLDA+DW: LightLDA, in which $\vect C_{d}$ is updated instantly, and $\vect C_{w}$ is updated per iteration.
\item LightLDA+DW+DD: LightLDA, in which both $\vect C_{d}$ and $\vect C_{w}$ are updated per iteration.
\item LightLDA+DW+DD+SP: LightLDA, in which both $\vect C_{d}$ and $\vect C_{w}$ are updated per iteration, using WarpLDA's $q^{\mbox{word}}$.
\item WarpLDA: WarpLDA, in which both $\vect C_{d}$ and $\vect C_{w}$ are updated per iteration, using WarpLDA's $q^{\mbox{word}}$.
\end{itemize}
Fig.~\ref{fig:impact-of-approximations} shows the results, where all the algorithms require roughly the same number of iterations to converge to a particular log-likelihood, with the same $M$. This result shows that the delayed update and simple proposal distributions of WarpLDA do not affect the convergence much. Notice that the previous results in Fig.~\ref{fig:convergence}, where WarpLDA requires more iterations than LightLDA to converge to a particular log-likelihood, may look different from this result. This is because LightLDA uses a larger $M$ in Fig.~\ref{fig:convergence}.

Our observation that delayed update does not affect convergence much seems to contradict with the previous observations for exact sampling algorithms~\cite{ahmed2012scalable}. A possible reason is that for MH-based algorithms (e.g., WarpLDA) the bottleneck of convergence is the efficiency of exploring $p(z_{dn})$ instead of the freshness of the counts; and thereby delayed update on the counts does not matter a lot.

We also analyze the impact of $M$ on the solution quality in Fig.~\ref{fig:sensitivity}. We can see that as $M$ gets larger WarpLDA converges faster. This is  probably because of the bias induced by the finite-length MH chain. To keep the storage overhead small, we stick to a small $M$ such as 1, 2 or 4, which leads to a sufficiently fast convergence.

\subsection{Scalability Results}
Finally, we demonstrate that WarpLDA can scale up to handle billion-scale documents on hundreds of machines. 

Fig.~\ref{fig:scalability}(a) shows the multi-threading speedup result for WarpLDA with $M=2$ and $K=10^3$ on the NYTimes dataset. The throughput for a single core, a single CPU (12 cores), and 2 CPUs (24 cores) are 6M, 53M, and 104M tokens per second, respectively. The speedup of the 24-core version against the single-core version is 17x, which is good for such a memory intensive task. The 2-CPU (24 cores) version is faster than the single CPU (12 cores) version by 1.96x, indicating that our NUMA strategy is successful.

Fig.~\ref{fig:scalability}(b) shows the multi-machine speedup result for WarpLDA with $M=1$ and $K=10^4$ on the PubMed corpus. The throughput for 16 machines is 13.5x faster than the single machine version, demonstrating the good scalability.

To show our capacity of learning large-scale topic models, we learned $K=10^6$ topics on the 639-million-document ClueWeb12 corpus, on 256 machines. The hyper-parameter $\beta$ is set to 0.001 for finer grained topics, $M$ is set to 1, and the number of iterations is set to 900. The convergence results are shown in Fig.~\ref{fig:scalability}(c). We can see that the run produces meaningful results in 5 hours.\footnote{The learned 1 million topics are available at \url{http://ml.cs.tsinghua.edu.cn/~jianfei/warplda.html}.} Fig.~\ref{fig:scalability}(d) shows the throughput is an unprecedentedly 11G tokens/s with 256 machines.



\section{Conclusions and Future Work}
We first analyze the memory efficiency of previous fast algorithms for LDA by the size of randomly accessed memory per-document, and conclude they are inefficient because of frequent random accesses to large matrices.  
We then propose WarpLDA, an efficient algorithm for LDA which only requires to randomly access vectors that fit in the L3 cache, while maintaining the $O(1)$ time complexity. WarpLDA builds on an MCEM algorithm that enables delayed updates to decouple the accesses to $\vect C_d$ and $\vect C_w$, by some carefully designed ordering of visiting tokens. To implement WarpLDA in a memory efficient and scalable way, we design and implement a framework that supports manipulating the rows and columns of a distributed sparse matrix.

Extensive empirical studies in a wide range of testing conditions demonstrate that WarpLDA is consistently 5-15x faster than the state-of-the-art MH-based algorithms, and is faster than state-of-the-art sparsity-aware algorithms in most settings. WarpLDA achieves an unprecedentedly 11G token per second throughput which allows to train billion-scale corpora in hours.

In the future, we plan to combine WarpLDA with other promising directions of scaling up LDA and apply it to more sophisticated topic models to learn various types of topic structures. 

\textbf{Stochastic learning: }
Stochastic algorithms explore the statistical redundancy of a given corpus, and estimate the statistics (e.g., gradient) of the whole corpus by a random subset. When the estimation has low variance, faster convergence is expected. Examples include stochastic variational inference (SVI)~\cite{hoffman2013stochastic}, streaming variational Bayes~\cite{broderick2013streaming}, and stochastic gradient Riemann Langevin dynamics (SGRLD)~\cite{patterson2013stochastic}. These methods can be combined with fast sampling algorithms, e.g. WarpLDA, to create fast stochastic algorithms, e.g., Bhadury et al. combined SGRLD with LightLDA to scale up dynamic topic models ~\cite{bhadury2016scaling}. 

\textbf{GPU accelerations: }
There are some works on GPU acceleration for LDA~\cite{canny2013bidmach,zhao2014same}. However, the algorithm they accelerate is $O(K)$. WarpLDA is a promising $O(1)$ option for GPU acceleration due to its single instruction multiple data (SIMD) nature. \footnote{We do not mention this in the main text, but readers can verify that the tokens in one document / word can be processed simultaneously without data race.}

\textbf{Non-conjugate topic models: } Compared to the vanilla LDA, non-conjugate topic models capture richer types of statistical structures, e.g., correlation of topics~\cite{blei2006correlated,chen2013scalable}, temporal dynamics of topics~\cite{blei2006dynamic,bhadury2016scaling}, or relationship to labels~\cite{zhu2013scalable,zhu2013bayesian}. These models typically have less sparsity structures to exploit, making sparsity-aware algorithms difficult to apply.  We can still apply the MH-based WarpLDA to these models as a fast sampler for topic assignments.

\section{Acknowledgments}
We thank Jinghui Yuan for the help on the experiments.
The work was supported by the National Basic Research Program (973 Program) of China (Nos. 2013CB329403, 2012CB316301), National NSF of China (Nos. 61322308, 61332007), Tsinghua TNList Lab Big Data Initiative, and Tsinghua Initiative Scientific Research Program (No. 20141080934).

\bibliography{example_paper}
\bibliographystyle{abbrv}

\begin{appendix}
\section{Full code of WarpLDA}
Alg.~\ref{alg:warplda} is the full pseudo-code of WarpLDA, with random positioning for sampling the document proposal and alias sampling for sampling the word proposal.
\begin{algorithm}[H]
\begin{minipage}{\linewidth}
    \caption{The WarpLDA algorithm, where Dice$(K)$ draws a sample from $\{0, \dots, K-1\}$. Alias table is used for sampling from $\vect C_w$ and positioning is used for sampling from $\vect C_d$.\label{alg:warplda}}
\label{residual}
\begin{algorithmic}
\State // Initialize
\State $z_{dn}\leftarrow \mbox{Dice}(K), \forall d, n$

\For {$iteration \leftarrow 1 \mbox{ \textbf{to} } I$}
\State // Word phase: $C_{wk}, \pi^{\mbox{doc}}, q^{\mbox{word}}$
\For {$w \leftarrow 1 \mbox{ \textbf{to} } V$}
\State // Compute $C_{wk}$ on the fly
\State $C_{wk} \leftarrow \sum_{n=1}^{L_w} \mathbb I(z_{wn}=k), k=1, \dots, K$
\State // Simulate $q^{\mbox{doc}}$ chain with samples from last iteration
\For {$n\leftarrow 1 \mbox{ \textbf{to} } L_w$}
    \For {$i \leftarrow 1 \mbox{ \textbf{to} } MH$}
        \State $s\leftarrow z_{wn}^{(i-1)}, t\leftarrow z_{wn}^{(i)}$
        \State $\pi \leftarrow \min \{1, \frac{C_{wt} + \beta}{C_{ws} + \beta}\frac{C_s + \bar\beta V}{C_t + \bar\beta}\}$
        \State $z_{wn} \leftarrow t$ with probability $\pi$
    \EndFor
\EndFor
\State // Update $C_{wk}$ 
\State $C_{wk} \leftarrow \sum_{n=1}^{L_w} \mathbb I(z_{wn}=k), k=1, \dots, K$
\State $urn\leftarrow \mbox{BuildAlias}(\vect C_w)$

\State // Draw samples from $q^{\mbox{word}}$
\For {$n\leftarrow 1 \mbox{ \textbf{to} } L_w$}
\For {$i \leftarrow 1 \mbox{ \textbf{to} } MH$}
\State $z_{wn}^{(i)} \leftarrow urn.\mbox{Draw}()$
\EndFor
\EndFor
\EndFor

\State // Document phase: $C_{dk}, \pi^{\mbox{word}}, q^{\mbox{doc}}$
\For {$d \leftarrow 1 \mbox{ \textbf{to} } D$}
\State // Compute $C_{dk}$ on the fly
\State $C_{dk} \leftarrow \sum_{n=1}^{L_d} \mathbb I(z_{dn}=k), k=1, \dots, K$
\State // Simulate $q^{\mbox{word}}$ chain with samples from last iteration
\For {$n\leftarrow 1 \mbox{ \textbf{to} } L_d$}
    \For {$i \leftarrow 1 \mbox{ \textbf{to} } MH$}
        \State $s\leftarrow z_{dn}^{(i-1)}, t\leftarrow z_{dn}^{(i)}$
        \State $\pi \leftarrow \min \{1, \frac{C_{dt} + \alpha_t}{C_{ds} + \alpha_s}\frac{C_s + \bar\beta}{C_t + \bar\beta}\}$
        \State $z_{dn} \leftarrow t$ with probability $\pi$
    \EndFor
\EndFor
\State // Draw samples from $q^{\mbox{doc}}$
\For {$n\leftarrow 1 \mbox{ \textbf{to} } L_d$}
\For {$i \leftarrow 1 \mbox{ \textbf{to} } MH$}
\State $z_{dn}^{(i)} \leftarrow \begin{cases}
z_{d, \mbox{Dice}(L_d)} & \mbox{with probability }\frac{L_d}{L_d + \bar\alpha} \\
\mbox{Dice}(K) & \mbox{otherwise}
\end{cases}$
\EndFor
\EndFor
\EndFor

\EndFor
\end{algorithmic}
\end{minipage}
\end{algorithm}
\end{appendix}

\end{document}